\documentclass[runningheads]{llncs}
\usepackage{graphicx}
\usepackage[section]{placeins}
\usepackage{amsmath}
\usepackage{float}
\usepackage{amssymb}
\usepackage{csquotes}
\usepackage{algorithm}
\usepackage{algorithmic}
\usepackage{graphicx}
\usepackage{subfigure}
\usepackage{caption}

\usepackage{listings}

\usepackage{color}
\definecolor{deepblue}{rgb}{0,0,0.5}
\definecolor{deepred}{rgb}{0.6,0,0}
\definecolor{deepgreen}{rgb}{0,0.5,0}
\DeclareFixedFont{\ttb}{T1}{txtt}{bx}{n}{11} 
\DeclareFixedFont{\ttm}{T1}{txtt}{m}{n}{11}  

\newcommand\pythonstyle{\lstset{
language=Python,
basicstyle=\ttm,
otherkeywords={self},             
keywordstyle=\ttb\color{deepblue},
emph={MyClass,__init__},          
emphstyle=\ttb\color{deepred},    
stringstyle=\color{deepgreen},
frame=tb,                         
showstringspaces=false            %
}}

\newcommand\pythoninline[1]{{\pythonstyle\lstinline!#1!}}

\lstnewenvironment{python}[1][]
{
\pythonstyle
\lstset{#1}
}
{}
\renewcommand{\and}{\\}

\begin{document}
\title{Anomaly detection with Wasserstein GAN}
%
%
\author{Ilyass Haloui\inst{1,2} \and Jayant Sen Gupta\inst{1} \and Vincent Feuillard\inst{1}  }
\authorrunning{I.Haloui}
%
\institute{Airbus AI Research, Toulouse France \\ 
\email{\{ilyass.haloui,jayant.sengupta,vincent.feuillard\}@airbus.com} \\
\and ISAE Supaero, Toulouse France \\
\email{ilyass.haloui@supaero.isae.fr}}
\maketitle              
\begin{abstract}
Generative adversarial networks are a class of generative algorithms that have been widely used to produce state-of-the-art samples. In this paper, we investigate GAN to perform anomaly detection on time series dataset. In order to achieve this goal, a bibliography is made focusing on theoretical properties of GAN and GAN used for anomaly detection. A Wasserstein GAN has been chosen to learn the representation of normal data distribution and a stacked encoder with the generator performs the anomaly detection. W-GAN with encoder seems to produce state of the art anomaly detection scores on MNIST dataset and we investigate its usage on multi-variate time series.      

\end{abstract}
\section{Context}
Airbus platforms are more and more connected. Newer aircraft are already equipped with data concentrators and connectivity to transmit sensor data collected during the whole flight to the ground, usually when at the gate. Older aircraft are currently retrofitted to install boxes that are able to do the same job. For military and heavy helicopters, HUMS (Health and Usage Monitoring System) also allows the collection of sensor data. Finally, satellites send regularly to the ground data collected from sensors, called telemetries.

Most of the time, the platforms behave normally and faults and failures are rare. To go beyond corrective and preventive maintenance, and anticipate future faults and failures, we have to look for any drift, any change in systems' behavior, in data that is normal almost all the time. Moreover, sensor data we collect is time series data. The problem we have is then anomaly detection in time series data.  

Generative Adversarial Networks are a popular technique to generate data from an original dataset, which can be of high dimension. In our case, generate new data could be useful to generate abnormal data when they are spotted, but we are more interested in the potential of such techniques to do anomaly detection for high dimensional data, as time series data we are dealing with. 

Our study will be illustrated by three use cases of increasing complexity, in order to well understand what we do with GAN. One use case is a simple two-dimensional distribution, the second is the classical image database MNIST and the third is a time series problem coming from UCI database. They will be further described in the next section. 

The third section will be dedicated to the description of GAN, classical and Wasserstein-GAN. The section will mix theoretical formalization and illustrations with the use cases.  

The fourth section will be dedicated to the usage of GAN for anomaly detection in the literature. It will help to show where our contribution, described in the last section, stands.

We will conclude and present potential way forward.

\section{Use Cases Description}
The goal of this article is to study the feasibility of performing anomaly detection with generative adversarial networks. We chose three different use cases to better understand how GAN work and their limitations in this field. The first one is a simple 2-dimensional multimodal dataset, the second one deals with high dimensional data such as MNIST images, the last one is a multivariate sensor dataset.

\subsection{2D Mixture of gaussians}\label{ssec:UC1}

\begin{figure}[!htb]
    \centering
    \includegraphics[width=0.6\textwidth]{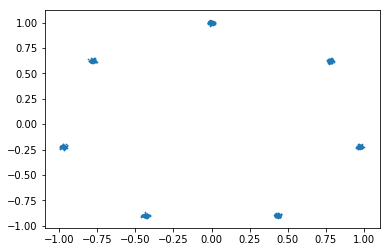}
    \caption{Dataset of 100 000 points sampled from a mixture of gaussians}
\label{fig:multi}
\end{figure}

First, we chose to study how GAN perform on a simple dataset. We try to learn a simple 2-dimensional distribution charaterized by a mixture of 7 gaussians with centers positioned on the unit circle. This use case have been used in several GAN papers investigating how well GAN perform on multi-modal datasets \cite{ref_WGANGP} \cite{Unrolled_GAN}. The dataset can be seen in Figure \ref{fig:multi}
    
In this context, an anomaly is considered as being every point outside the level set of the gaussian centers. 

\subsection{MNIST digits}\label{ssec:UC2}

\begin{figure}[!htb]
    \centering
    \includegraphics[width=0.7\textwidth]{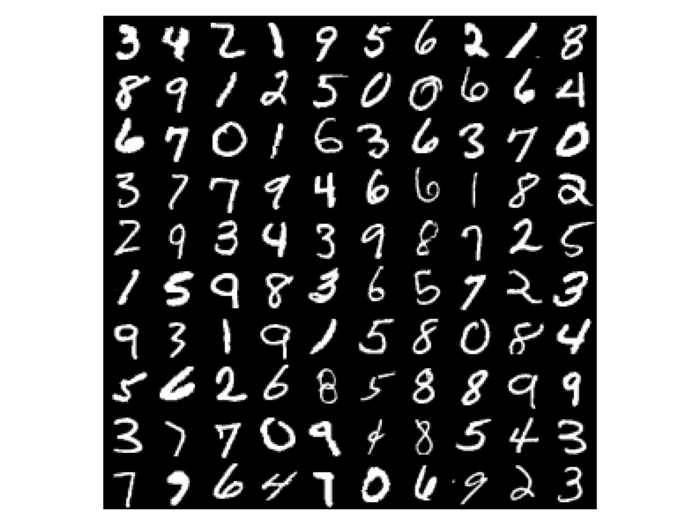}
    \caption{MNIST Digits}
\label{fig:mnist}
\end{figure}

We also study the well known MNIST digits dataset shown in Figure \ref{fig:mnist}. The goal is to build a GAN that generates good visual representations of digits from 0 to 9. The dataset has 60 000 digits for training and 10 000 for testing. Each digit is an image of 28x28 pixels and represents the first step towards the study of high dimensional data.  We perform anomaly detection considering one class of digits as being abnomal and train the GAN on the other digits of the training dataset. Testing is performed on the test dataset plus all the anomalous samples. This procedure have been used in \cite{EffiGAN} and \cite{VAE} and to investigate the performance of GAN in the field of anomaly detection. To compare results, we compute the Area Under Precision Recall Curve (AUPRC) for each digit.

\subsection{UCI HAR}
This dataset is constituted of mutivariate time series of smartphone sensor data. The movement data recorded was the x, y, and z accelerometer data (linear acceleration) and gyroscopic data (angular velocity) from the smart phone. Observations were recorded at 50 Hz (i.e. 50 data points per second). There are three main signal types in the raw data: total acceleration, body acceleration, and body gyroscope. Each has three axes of data. This means that there are a total of nine variables for each time step. \\
The data is acquired in 6 labeled different situations, the subject carrying the smartphone is whether walking, standing, walking upstairs, walking downstairs or laying. As for the MNIST dataset, we train the GAN and perform anomaly detection considering one class as being abnormal and the others as being normal. Figure \ref{fig:multiTS} shows an example of time series acquisition of 9 sensors for a walking individual on 128 time steps:   
\begin{figure}[!htb]
    \centering
    \includegraphics[width=0.5\textwidth]{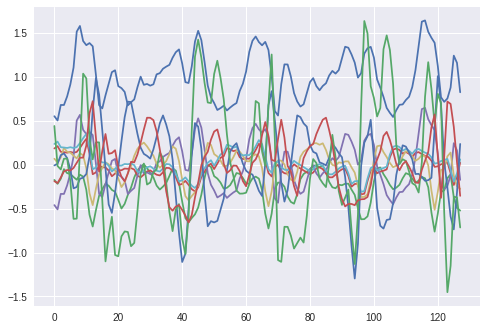}
    \caption{Example of time series recorded while walking }
\label{fig:multiTS}
\end{figure}

\section{Generative Adversarial Networks formulations}

\subsection{GAN description}
\begin{figure}[!htb]
    \centering
    \includegraphics[width=0.7\textwidth]{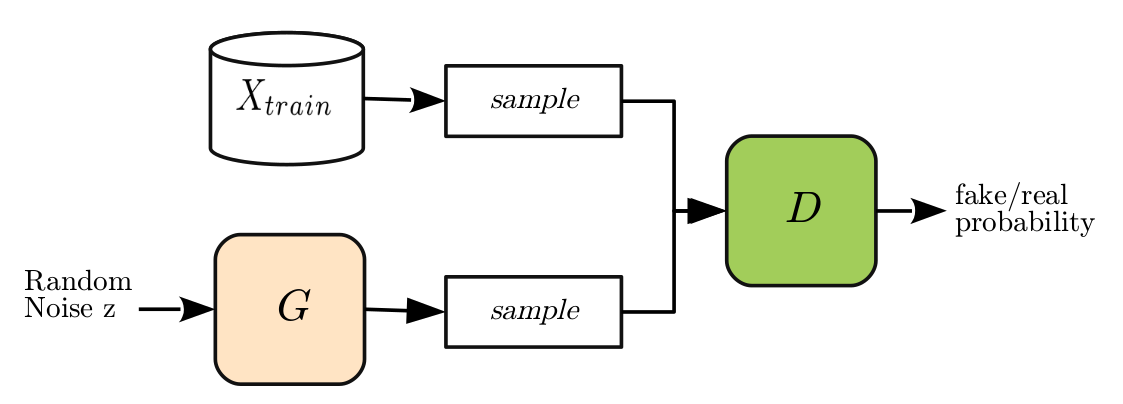}
    \caption{Generative Adversarial Networks}
\label{fig:GAN}
\end{figure}

Generative Adversarial networks have been proposed by \cite{GAN_goodfellow} as a new framework for estimating generative models via an adversarial process. Two models are trained simultaneously: a generative model G that captures the data distribution and a discriminative model D that estimates the probability that a sample came from the training data rather than from G. Figure \ref{fig:GAN} describes the training procedure.\\
The generative model can be thought of as analogous to a team of counterfeiters, trying to produce fake currency and use it without detection, while the discriminative model is analogous to the police, trying to detect the counterfeit currency. Competition in this game drives both teams to improve their methods until the counterfeits are indistinguishable from the genuine
articles. \\
Even if this framework can be applied to many kinds of models, the approach is mostly performed with two neural networks for G and D. To learn the generator’s distribution $\mathbb{P}_g$ over data x, a prior is defined on input noise variables $p_z(z)$, and its mapping to data space as $G(z; \theta_g)$. The discriminator $D(\mathbf{x}; \theta_d)$ is defined to output a single scalar. $D(\mathbf{x})$ represents the probability that $\mathbf{x}$ came from the data rather than the generator. \\
G and D play the following two-player minimax game with value function $V (G, D)$:
\begin{equation}
    \min_{G} \max_{D} V (D,G) = \mathbb{E}_{\mathbf{x} \sim p_{data}(\mathbf{x})} \log [D(\mathbf{x})] + \mathbb{E}_{\mathbf{z} \sim p_{\mathbf{z}}(\mathbf{z})} \log [1 - D(G(\mathbf{z}))]
\end{equation}

We train D to maximize the probability of assigning the correct label to both training examples and samples from G. We simultaneously train G to minimize $\log [1 - D(G(\mathbf{z}))$.The algorithm proposed by \cite{Goodfellow_cite} is presented in Algo.\ref{alg:algo_gan}.

\begin{algorithm}
\caption{Minibatch stochastic gradient descent training of generative adversarial nets. The number of steps to apply to the discriminator, k, is a hyperparameter. k = 1 in most of GAN experiments.}
\begin{algorithmic}[H]

\FOR {number of training iterations }
\FOR{k steps}
\STATE Sample $\{x^{(i)}\}_{i = 1}^{i=m}$ $\sim$ $\mathbb{P}_r$ a batch from the real data.
\STATE Sample $\{z^{(i)}\}_{i = 1}^{i=m}$ $\sim$ $p(z)$ a batch of prior samples.

\STATE Update the discriminator by ascending its stochastic gradient: $\nabla_{\theta_d} \frac{1}{m} \sum_{i=1}^{m}( \log (D(x^{(i)}) + (1 - \log (D(G(z^{(i)})))$

\ENDFOR
\STATE Sample $\{z^{(i)}\}_{i = 1}^{i=m}$ $\sim$ $p(z)$ a batch of prior samples
\STATE Update the generator by ascending its stochastic gradient: $\nabla_{\theta_g} \frac{1}{m} \sum_{i=1}^{m} (1 - \log D(G(z^{(i)})))$

\ENDFOR

\STATE The gradient-based updates can use any standard gradient-based learning rule
\end{algorithmic}

\label{alg:algo_gan}
\end{algorithm}

During the past four years, GAN have been successfully used to produce state of the art samples in image generation. In the next subsection, we investigate in more details their mathematical properties.  

\subsection{GAN properties}
GAN theoretical properties have been discussed in \cite{GAN_properties} and \cite{GAN_goodfellow} , and is still subject to fundamental research.
In this section, we describe some optimality characteristics of GAN convergence.\\ 
  
\subsubsection{Notations:}
\begin{itemize}
    \item[$\bullet$]   Let us consider the data space $\mathcal{X}$  as being a d-dimensional borelian subset of $\mathbb{R}^d$, dominated by a mesure $\mu$, and $(X_1 , .. , X_n) \sim \mathbb{P}_r$ on $\mathcal{X}$. Each sample $X_i$ is a random variable that takes values on a high dimensional space $(d >> 1)$.
    \item[$\bullet$]  The Generators are considered to be part of a parametric family of functions $\mathcal{G}$ = $\left\{ G_{\theta} \right\}_{\theta \in \Theta} $, where $G_{\theta} : \mathbb{R}^{d'} \mapsto \mathcal{X}$. The latent space $\mathbb{R}^{d'}$ is usually a much lower dimensional space than $\mathcal{X}$ $(d' << d)$.  By having $Z \sim \mathcal{N}(0,1)^{d'}$, we can define $G_{\theta}(Z) \sim \mathbb{P}_{\theta} $, where $\mathbb{P}_{\theta}  \in \left\{\mathbb{P}_{\theta}  \right\}_{\theta \in \Theta} $ is a potential candidate to represent $\mathbb{P}_{r} $.
    \item[$\bullet$]  The Discriminators are also part of a parametric family of functions $\mathcal{D}$ = $\left\{ D_{\alpha} \right\}_{\alpha \in \Lambda} $, where $D_{\alpha }$ : $ \mathcal{X} \mapsto \lbrack 0 , 1 \rbrack $.
\end{itemize}

To this point, we assume that the generator and the discriminator can actually be simulated by any parametric model (not necessarily neural networks). Also, we do not assume that the target distribution $\mathbb{P}_r$ correspond to a specific parameterization $\theta$, since we only have samples from $\mathbb{P}_r$. The goal of this section is to study the links and deviations between these two distributions at optimality. 
\\ The standard objective function in GAN training is: 
\begin{equation}
    \underset{\theta}{\text{inf}} \ \underset{\alpha }{\text{sup}} \prod_{i =1}^n {D_{\alpha}(X_i)}\prod_{i =1}^n {(1- D_{\alpha}\circ G_{\theta}(Z_i))}
\end{equation}
We want to describe the following errors:
\begin{enumerate}
    \item  The error of the optimal distribution $\mathbb{P}_{\theta}$ due to the parameterization of the generator in $\theta$
    \item  The error of the optimal distribution $\mathbb{P}_{\theta}$ due to the parameterization of the discriminator in $\alpha$
    \item  The generalization error of the model due to the sample size $n$
\end{enumerate}

\begin{figure}[!htb]
    \centering
    \includegraphics[width=0.85\textwidth]{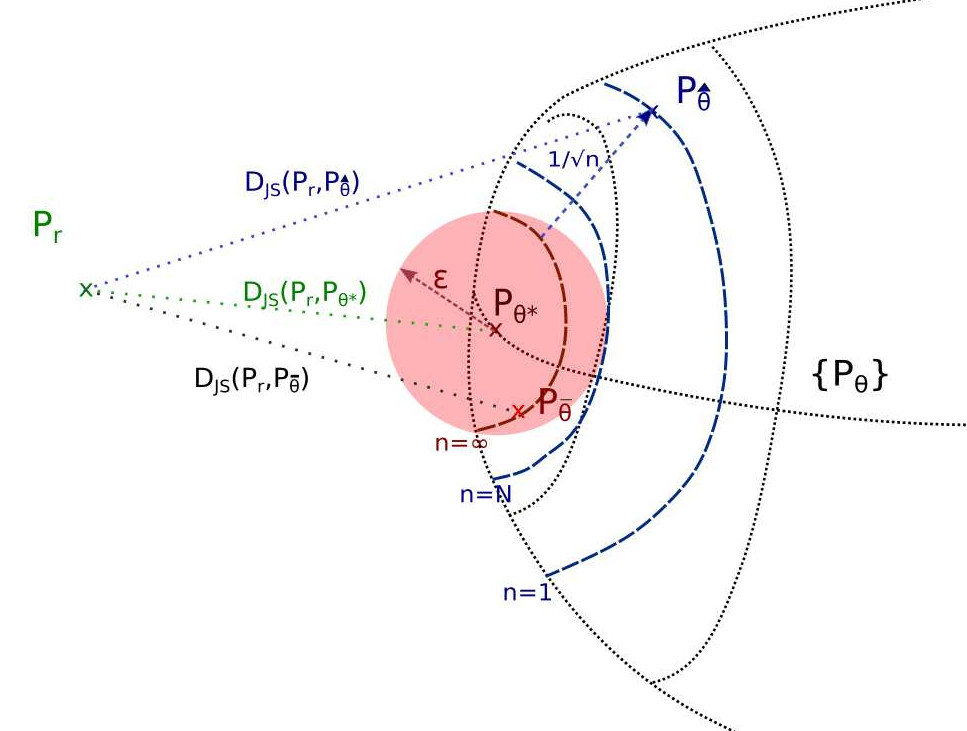}
    \caption{Summary of origins of errors on $\mathbb{P}_r$ approximation}
\label{fig:errors_gan}
\end{figure}

To compare probability distributions, we use the Jensen-Shannon divergence as in \cite{GAN_properties}. Figure \ref{fig:errors_gan} helps keep track of these error.

\subsubsection{Error in generator parameterization} \hfill\\
To quantify the first error, we rewrite the objective relatively to $\mathbb{P}_r$ only:\\
\begin{equation}
 \underset{\theta}{\text{inf}} \underset{D \in \mathcal{X} \mapsto [0 , 1] }{\text{sup}}
    L(\theta, D) \triangleq \int ln(D)\mathbb{P}_r d\mu + \int ln(1 - D)\mathbb{P}_{\theta} d\mu 
\end{equation}

If L is bounded, then:
    \begin{theorem}
   Optimal discriminator
     \begin{center}
         $\exists! D^*_{\theta}$ such that $D^*_{\theta} = \underset{D \in \mathcal{D_{\infty}}  }{\text{argmax }} L(\theta,D)$ \\
        
    \end{center}

    \end{theorem}
    The optimal discriminator is :
    \begin{equation}
        D^*_{\theta} \triangleq \frac{\mathbb{P}_r}{\mathbb{P}_r+\mathbb{P}_{\theta}}
    \end{equation}
    And the supremum of L is achieved : 
    \begin{equation}
        \underset{D \in \mathcal{D_{\infty}}  }{\text{sup }} L(\theta,D) = L(\theta,D^*_{\theta}) = 2 D_{JS}(\mathbb{P}_r ,\mathbb{P}_{\theta}) - ln(4) , \forall \theta \in \Theta
    \end{equation}
Where $D_{JS}$ denotes for Jensen-Shannon divergence

\begin{theorem} 
Optimal parameterized generator:
\begin{center}
      If $\exists$ m,M such that : $m \leq \mathbb{P}_r\leq M$ and $\forall \theta \in \Theta$, $  \mathbb{P}_{\theta} \leq M$, then:\\
$\exists! \theta^* \in \Theta $ such that $\theta^* = \underset{\theta \in \Theta  }{\text{argmin }} L(\theta,D^*_{\theta})$
$\Longleftrightarrow$ 
$\theta^* = \underset{\theta \in \Theta  }{\text{argmin }} D_{JS}(\mathbb{P}_r,\mathbb{P}_{\theta})$
\end{center}
  
\end{theorem}

The above theorem demonstrated in \cite{GAN_properties} shows the existence of an optimal parametric generator independently from the discriminator parameterization and the sample size $n$. Figure \ref{fig:errors_gan} illustrates in green the Jensen-Shannon divergence between the true distribution $\mathbb{P}_r$ and this optimal generator distribution $\mathbb{P}_{\theta^*}$.  In order to perform optimization in practice, a parameterization of the discriminator is also needed and results in an additional approximation error on $\theta$.

\subsubsection{Resulting error in discriminator parameterization}\hfill\\
In practice, the set of all possible discriminators is only a subset of $\mathcal{D}_{\infty}=\{ \mathcal{X} \mapsto [0 , 1] \}$. $\mathcal{D}$ = $\left\{ D_{\alpha} \right\}_{\alpha \in \Lambda} $, where $D_{\alpha }$ : $ \mathcal{X} \mapsto \lbrack 0 , 1 \rbrack $ is parameterized in $\alpha$\\
We are looking for $\overline{\theta} \in \Theta $ such that:
\begin{equation}
    \underset{D \in \mathcal{D} }{\sup} L(\overline{\theta},D) \leq \underset{D \in \mathcal{D} }{\sup} L(\theta,D), \forall \theta \in \Theta
\end{equation}

\begin{theorem}
Error on discriminator parameterization:
\begin{center}
    If $(H_0)$ : $\exists \underline{t} \in \rbrack 0 , 1/2 \rbrack $ such that $\min(D^*_{\theta}, 1 - D^*_{\theta}) \geq \underline{t} $, ie \\
 $\displaystyle \frac{\underline{t}}{1 - \underline{t}} \leq  \displaystyle \mathbb{P}_{\theta} \leq \frac{1- \underline{t}}{\underline{t}}$\\
If $(H_{\epsilon}): \exists  \epsilon \in (0,\underline{t}) $ and $D \in \mathcal{D}$ such that: $\|D - D_{\overline{\theta}}^* \| \leq \epsilon  $\\Then:\\
$\exists c_{\underline{t}} $ such that $0 \leq D_{JS}(\mathbb{P}_r,\mathbb{P}_{\overline{\theta}}) - D_{JS}(\mathbb{P}_r,\mathbb{P}_{\theta^*}) \leq c_{\underline{t}} \epsilon$
\end{center}
\end{theorem}

This second theorem shows that under certain conditions, the parameterization of the discriminator results in an error $\epsilon$ on the generator distribution $\mathbb{P}_{\overline{\theta}}$. Figure \ref{fig:errors_gan} shows in black the resulting Jensen-Shannon divergence between the true distribution $\mathbb{P}_{r}$ and the optimal generator distribution after parameterization of the discriminator $\mathbb{P}_{\overline{\theta}}$

\subsubsection{Sample size and distribution convergence}\hfill\\
We are interested on evaluating the error between the true distribution of the dataset $\mathbb{P}_r$ and the generator distribution $\mathbb{P}_{\theta}$, knowing the parameterization of the generator in $\theta$, the parameterization of the discriminator in $\alpha$ and the sample size $n$ of the dataset.  \\
To do so, we go back to the original GAN formulation we want to minimize w.r.t $\theta$ and maximize w.r.t. $\alpha$: 
\begin{equation}
     \hat{L}(\theta , \alpha) \triangleq \sum_i ln(D_{\alpha}(X_i) + \sum_i ln(1-D_{\alpha}\circ G_{\theta}(Z_i)) 
\end{equation}
Let's consider the following theorem from \cite{GAN_properties}:
\begin{theorem}
Suppose $(H_{reg})$:
\begin{center}
    $(H_{D})$ : $\exists k \in \rbrack 0 , 1/2 \rbrack $: $k \leq D_{\alpha} \leq 1-k $ where $(x,\alpha) \mapsto D_{\alpha}(x)$ is $C^1$ and bounded \\

$(H_{G})$ : $\forall z \in \mathbb{R}^{d'} $ , e  $\theta \mapsto G_{\theta}(z)$ is $C^1$ and bounded \\
$(H_{p})$ : $\forall x \in E $, on a $\theta \mapsto \mathbb{P}_{\theta}(x)$ is $C^1$ and bounded \\ 
\end{center}
Suppose $(H_{\epsilon})$:
\begin{center}
         $\exists \epsilon \in \rbrack 0 , \underline{t} \rbrack $ such that 
         $\forall \theta \in \Theta , \exists D \in \mathrm{D} $ such that $\|D - D_{\overline{\theta}}^* \|_{\infty} \leq \epsilon$\\
\end{center}
Then we have:
\begin{center}
        $\mathbb{E}D_{JS}(\mathbb{P}_r,\mathbb{P}_{\hat{\theta}}) - D_{JS}(\mathbb{P}_r,\mathbb{P}_{\theta^*}) = O(\epsilon^2 + \frac{1}{\sqrt{n}})  $\\
\end{center}
\end{theorem}
The above theorem tells us that under some reasonable conditions, the error on the generator convergences towards the true distribution $\mathbb{P}_r$ is proportional to $\frac{1}{\sqrt{n}}$, which is illustrated in blue in Figure \ref{fig:errors_gan}. \\Actually, if the generator is $C^2$ w.r.t $\theta$ with bounded derivatives, the discriminator is bounded and $C^2$ w.r.t $\alpha$ and $x$ with bounded derivatives, $\mathbb{P}_{\theta}$ is $C^2$ w.r.t $\theta$ also with bounded derivatives, then the larger the sample size the lower the generalization error and we can show that  $ \hat{\theta} \rightarrow \overline{\theta}$ and $ \hat{\alpha} \rightarrow \overline{\alpha}$ \cite{GAN_properties}.

\subsubsection{Summary}\hfill\\
In a nutshell, the Jensen-Shannon divergence seems to be a good metric to compare distributions during GAN training. Also, the parameterization of the discriminator part of the GAN results in an incompressible error that can only be mitigated by the choice of the discriminator. Finally, the larger sample size the better the convergence towards the data distribution. All these errors are described in Figure \ref{fig:errors_gan} where the space corresponds to all  functions defined in $\mathcal{X} \longrightarrow [0 , 1]$. In addition, the results demonstrated in \cite{GAN_properties} are independent from the chosen parameterization. Most GAN training procedures use Neural Networks as parameterized functions which involves fine tuning of hyperparameters in order to achieve the desired convergence. 

\subsubsection{Illustration:}\hfill\\
To apply Generative Adversarial Networks, we use two fully connected neural networks as the generator and the discriminator in order to learn the distribution of the first use case. Both networks have two hidden layers of 128 neurons each. The latent space have a dimension of two and the noise is sampled from a gaussian distribution with 0 mean and 1 variance. We sample 100 000 2-dimensional points from the mixture of gaussians and constitute our dataset. 
\begin{figure}
    \centering
    \includegraphics[width=.5\textwidth]{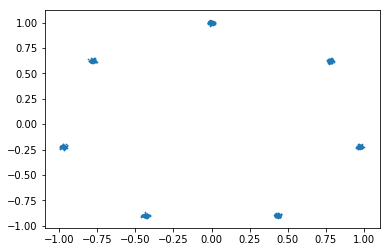}
    \caption{Dataset sampled from mixture of gaussians}
\label{fig:my_label}
\end{figure}
We perform standard learning procedure on this dataset and obtain a loss curve for the discriminator displayed in Figure \ref{fig:loss_toy}. We also decide to plot the output of the generator at each training epoch:
\begin{figure}[!htb]
    \centering
    \includegraphics[width=.5\textwidth]{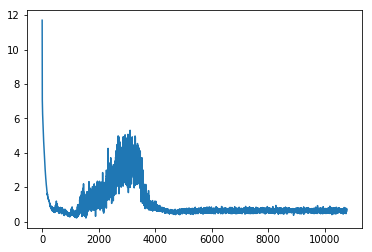}
    \caption{Loss curve of the discriminator in GAN training}
\label{fig:loss_toy}
\end{figure}
The training loss is difficult to interpret since the global minimum for GAN training involves the generator and the discriminator. Yet it is hard to deduce convergence from loss curves.

\begin{figure}[!htb]
\centering
\subfigure{\includegraphics[width=.24\linewidth]{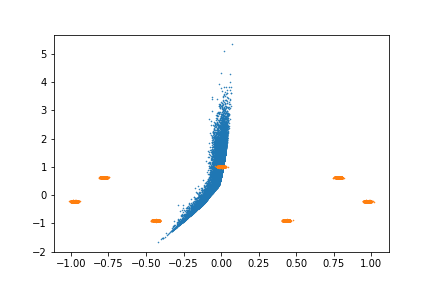}}
\subfigure{\includegraphics[width=.24\linewidth]{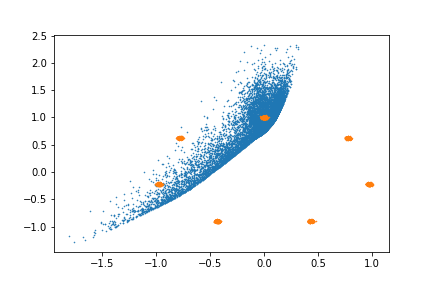}}
\subfigure{\includegraphics[width=.24\linewidth]{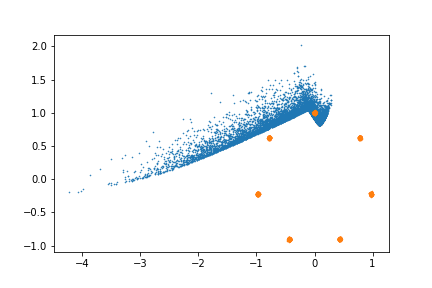}}
\subfigure{\includegraphics[width=.24\linewidth]{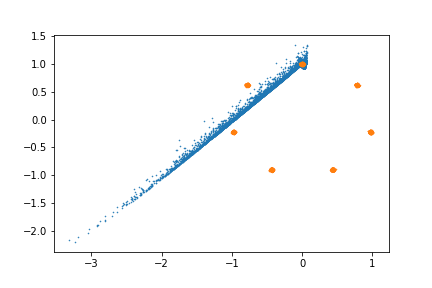}}
\subfigure{\includegraphics[width=.24\linewidth]{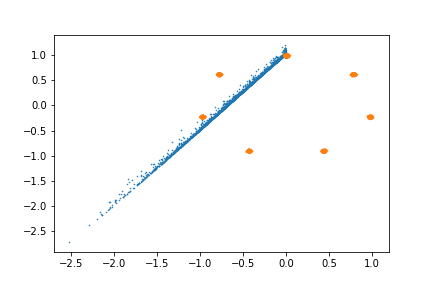}}
\subfigure{\includegraphics[width=.24\linewidth]{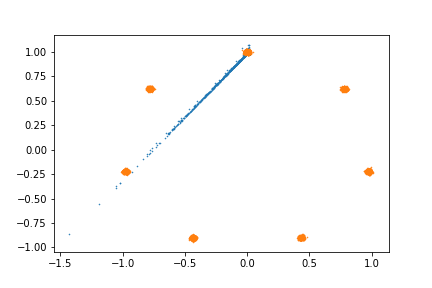}}
\subfigure{\includegraphics[width=.24\linewidth]{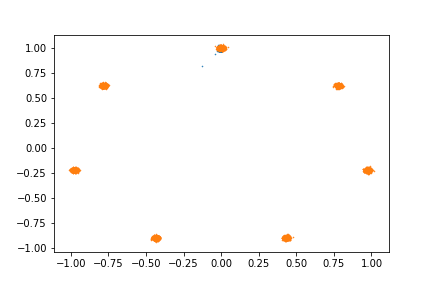}}
\subfigure{\includegraphics[width=.24\linewidth]{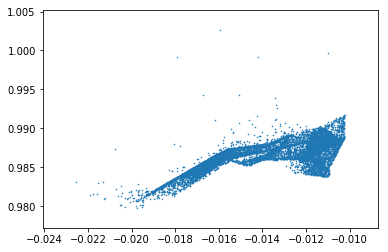}}
%
\caption{Output of the generator for the first 8 epochs (from top to bottom, left to right). The GAN collapses towards one cluster of the distribution of coordinates (0,1). The last figure have different scale to show the position of the
generated points}
\label{fig:Collapse}
\end{figure}

It is clear that the learning has not been performed as expected. This is due to a famous problem that occurs during GAN training: the generator collapses and only produces samples similar to a small subset of the distribution.  This can be understood from the objective function: if the generator starts producing samples coming from one mode that fool the discriminator, then it is encouraged to generate only similar samples and thus forget the other "modes" that exist in the dataset. This is a major problem if we want to perform anomaly detection with GAN: it is important to learn all the distribution of normal data and not only a subset of this distribution. 

\subsection{Wasserstein Generative Adversarial Network}
As it was discussed in the previous sections, GAN usually suffer from mode collapse problem. Instead of learning a good representation of the data, the generator only learns to reproduce a small fraction of the variability of the dataset. This is due to GAN training procedure: since the generator is rewarded if it produces good realistic samples (by fooling the discriminator), it is not encouraged to produce other samples that might be not as good for the discriminator as the ones already found. Nevertheless, these other samples might help capture other existing "modes" in the dataset. To address this problem, a recent investigation focused on directly learning the distribution of the dataset using optimal transport theory instead of focusing on samples in a zero-sum game. \\ Moreover, avoiding mode collapse is a necessity if one needs to perform anomaly detection based on GAN: if the generator is not able to reproduce samples from all the data distribution, then missing modes can be interpreted as anomalies which increases the rate of false positives.

\subsubsection{Optimal Transport and generative modelling}\hfill\\
 Wasserstein Generative Adversarial Networks (W-GAN) \cite{ref_WGAN} were first introduced as a solution to the mode collapse problem, focusing on a specific objective function based on the Wasserstein distance (sometimes called Earth Mover Distance in the discrete case) between the generator distribution of samples $\mathbb{P}_{\theta}$ and the true distribution $\mathbb{P}_r$. Instead of learning directly from the samples, the GAN objective function tries to learn the underlying distribution $\mathbb{P}_r$ of the dataset. \\  

\par The Wasserstein distance effectively compares two distribution by considering the expected value of the distance between samples from the distributions two by two. This process makes the link with optimal transport theory. The original Kantorovich formulation for the Wasserstein distance in optimal transport theory is \cite{Villani}:
\begin{equation}
    W(\mathbb{P}_r,\mathbb{P}_{\theta}) =\underset{\gamma \in \Pi(\mathbb{P}_r,\mathbb{P}_g)}{\inf} \mathbb{E}_{(x,y) \sim \gamma} [\|x-y\|]
\end{equation}
 Where $\Pi(\mathbb{P}_r,\mathbb{P}_{\theta})$ denotes the set of all joint distributions $\gamma(x,y)$ with $\mathbb{P}_r$ and $\mathbb{P}_{\theta}$ as their marginals.  
\begin{displayquote}
Intuitively, $\gamma(x,y)$ indicates how much “mass”
must be transported from $x$ to $y$ in order to transform the distributions $\mathbb{P}_{r}$ into the distribution $\mathbb{P}_{\theta}$. The Earth Mover distance then is the “cost” of the optimal transport plan. \cite{ref_WGAN}
\end{displayquote}

From the generative point of view, the generator is parameterized by $\theta$ and its underlying distribution is $\mathbb{P}_{\theta}$. The goal is to build a distribution $\mathbb{P}_{\theta}$ that is the closest to the data distribution $\mathbb{P}_r$. \\ Continuity of the Wasserstein distance is necessary and demonstrated in the original paper and given by the following theorem : 
\begin{theorem}[Continuity]
Let $g_{\theta}$ be any feedforward neural network parameterized by $\theta$, and $p(z)$ a prior over z such that $\mathbb{E}_{z \sim p(z)} [\|z\|] < \infty$ (e.g. Gaussian, Uniform, etc .).\\
Then g is continuous in $\theta$ and therefore $W(\mathbb{P}_r,\mathbb{P}_{\theta})$  is continuous everywhere
and differentiable almost everywhere. \\
\end{theorem}
 The continuity statement is not satisfied by classical distances or divergences such as the Jensen-Shannon divergence, Kullback-Leibler divergence or total variation distance, which motivates the usage of the Wasserstein distance among others. 
Moreover, the Wasserstein distance guarantees the convergence towards 0 when the distributions become increasingly closer. This property allows having meaningful convergence metrics. \\ 
\par
Unfortunately, the calculation of the Wasserstein distance in its original form is intractable. Instead, its dual form given by the Kantorovich-Rubinstein duality allows nicer optimization properties: 
\begin{equation}
    W(\mathbb{P}_r,\mathbb{P}_{\theta}) = \underset{\|f\|_ {L<1}}{\sup}\mathbb{E}_{x \sim \mathbb{P}_r}[f(x)] - \mathbb{E}_{x \sim \mathbb{P}_{\theta}}[f(x)]  
\end{equation}

The optimization is performed over the set of all 1-Lipschitz functions $ \{ f: x \in \mathcal{X} \longmapsto f(x) \in \mathbb{R} \} $.  Using this formulation, the intuition is to consider a neural network that will play the role of this variable $f$ called the "critic" which is analog to the discriminator in the classical GAN formulation. 

\subsubsection{Wasserstein GAN formulation}\hfill\\
Let us consider the Wasserstein distance $W(\mathbb{P}_r,\mathbb{P}_{\theta})$ between the data distribution $\mathbb{P}_r$, and the distribution of the generated samples of the generator $\mathbb{P}_{\theta}$. The following theorem proves that the objective function is well defined and admits gradient formulation with respect to the generator parameters.
\begin{theorem}
Let $\mathbb{P}_r$ be any distribution. Let $\mathbb{P}_{\theta}$ be the distribution of $g_{\theta}(Z)$ with Z a random variable with density p and $g_{\theta}$  continuous in $\theta$.  Then, there is a solution $f : \mathcal{X} \xrightarrow{} \mathbb{R} $ to the problem :
\begin{center}
     $W(\mathbb{P}_r,\mathbb{P}_{\theta}) = \underset{\|f\|_ {L<1}}{\max}\mathbb{E}_{x \sim \mathbb{P}_r}[f(x)] - \mathbb{E}_{x \sim \mathbb{P}_{\theta}}[f(x)]  $
\end{center}
and we have :
\begin{center}
    $\nabla_{\theta}W(\mathbb{P}_r,\mathbb{P}_{\theta}) = - \mathbb{E}_{z \sim p(z)}[\nabla_{\theta} f(g_{\theta}(z))] $
\end{center}
\end{theorem}
If we consider a set of parameterized family of functions $\{f_w\}_{w \in \mathcal{W}}$ where the parameters are in a compact space $\mathcal{W}$, then all these functions are K-Lipchitz for some contraint K instead of being in the set of all 1-Lipschitz functions defined in $\mathcal{X} \longmapsto f(x) \in \mathbb{R}$. According to \cite{ref_WGAN}, this only yields to the calculation of the Wasserstein distance up to a multiplicative constant K. As a result, if we suppose that the supremum is attained for some $w \in  \mathcal{W} $, the new formulation of the Wasserstein distance  becomes:
\begin{equation}
    W(\mathbb{P}_r,\mathbb{P}_{\theta}) = \underset{w \in \mathcal{W}}{\max }  \mathbb{ E}_{x \sim \mathbb{P}_r}[f_w(x)] - \mathbb{E}_{z \sim p(z)}[f_w(g_{\theta}(z))] 
\end{equation}

To approximate this, we train a neural network parameterized with weights $w$ lying in a compact space
$\mathcal{W}$ and then back-propagate on $\mathbb{E}_{z \sim p(z)}[\nabla_{\theta} f_w(g_{\theta}(z))]$. Saying that $\mathcal{W}$ is compact means that all functions $f_w$ has to be lipschitz according to some constrain K. To build such functions, we can ensure that all weights must lie on a compact space. The first solution described in \cite{ref_WGAN} is to clip weights of the "critic" f in a fixed box $\mathcal{W} =[-0.01,0.01]$ which leads to convergence of training despite the restrictive solution. The original authors of the Wasserstein GAN states in \cite{ref_WGAN} that:
\begin{displayquote}
 Weight clipping is a clearly terrible way to enforce a Lipschitz constraint. If the clipping parameter is large, then it can take a long time for any weights to reach their limit, thereby making it harder to train the critic till optimality. If the clipping is small, this can easily lead to vanishing gradients when the number of layers is big, or batch normalization is not used (such as in RNNs)
\end{displayquote}

Using this technique, no collapse mode is observed during test phase as it can be shown in Figure \ref{fig:WGAN_toy}.

\subsection{Improved Wasserstein GAN algorithm}
 Instead of clipping the weights, some authors suggest to use another technique to ensure the respect of the Lipschitz constraint during training. As suggested in \cite{ref_WGANGP}, we can use the equivalence between Lipschitz constraint in the differentiable case and the norm of the gradients that has to be equal to the Lipschitz constraint. The objective function defined helps achieve the lipschitz constraint by penalizing the objective function with an additional term forcing the gradient to be smaller than 1:

\begin{equation}
    W(\mathbb{P}_r,\mathbb{P}_{\theta}) =\max   \mathbb{ E}_{x \sim \mathbb{P}_r}[f_w(x)] - \mathbb{E}_{z \sim p(z)}[f_w(g_{\theta}(z))] + \lambda .\mathbb{ E}_{x \sim \mathbb{P}_{tot}} [( \| \nabla_x f (x) \|_2 - 1 )^2]
\end{equation}

In practice, the penalization coefficient has to be well defined in order to achieve convergence. Also, the gradient calculation cannot be achieved everywhere in the state space $\mathcal{X}$. Instead, the gradient is calculated in each sample drawn from the data distribution and the generator distribution. Every point sampled from the straight line between the true samples and the generated samples is forced to have a gradient of 1.  

\begin{algorithm}
\caption{WGAN proposed algorithm. All experiments in the paper used
the default values $\alpha$ = 0.0001, $\lambda$ = 10, m = 64, $n_{critic}$ = 5, $\beta_1$ = 0.5, $\beta_2$ = 0.9}
\label{alg:WGAN}
\begin{algorithmic}[H]
\REQUIRE $\alpha$, the learning rate. c, the clipping parameter. m, the batch size. $n_{critic}$, the number of iterations of the critic per generator iteration.
\REQUIRE $w_0$, initial critic parameters. $\theta_0$, initial generator’s parameters.
\WHILE {$\theta$ has not converged}
\FOR {$t = 0 ,..., n_{critic}$}
\STATE Sample $\{x^{(i)}\}_{i = 1}^{i=m}$ $\sim$ $\mathbb{P}_r$ a batch from the real data.
\STATE Sample $\{z^{(i)}\}_{i = 1}^{i=m}$ $\sim$ $p(z)$ a batch of prior samples.
\STATE Sample $\epsilon$ $\sim$ $U[0,1]$
\STATE $\hat{x}^{(i)}$ $\longleftarrow$ $\epsilon x^{(i)} + (1-\epsilon).g_{\theta}(z^{(i)})$
\STATE $L_w^{(i)}$ $\longleftarrow$ $ [ f_w(x^{(i)}) -  f_w(g_{\theta}(z^{(i)})) + \lambda.[( \| \nabla_x f_w (\hat{x}^{(i)}) \|_2 - 1 )^2]]$

\STATE $w \longleftarrow $ Adam($\nabla_{w} \frac{1}{m}\sum\limits_{i=1}^m L_w,w,\alpha,\beta_1,\beta_2$)

\ENDFOR
\STATE Sample $\{z^{(i)}\}_{i = 1}^{i=m}$ $\sim$ $p(z)$ a batch of prior samples.
    \STATE $\theta \longleftarrow $Adam($ - \nabla_{\theta} \frac{1}{m} \sum\limits_{i=1}^m f_w(g_{\theta}(z^{(i)})), \theta, \alpha, \beta_1, \beta_2$)

\ENDWHILE
\end{algorithmic}
\end{algorithm}
Algorithm \ref{alg:WGAN} is the Wasserstein GAN algorithm with gradient penalty used in \cite{ref_WGANGP}, the hyperparameters needed are the parameters $\alpha, \beta_1$ and $ \beta_2$ used for the optimizer, $n_{critic}$ which corresponds to the ratio between the number of training steps of the discriminator (or critic) over the the generator, the gradient penalty coefficient $\lambda $ enforcing the Lipschitz constraint and the batch size $m$. 

\subsubsection{Illustration}
\begin{figure}[!htbp]
\centering

\subfigure{\includegraphics[width=.45\linewidth]{multigaussian.png}}
\subfigure{\includegraphics[width=.45\linewidth]{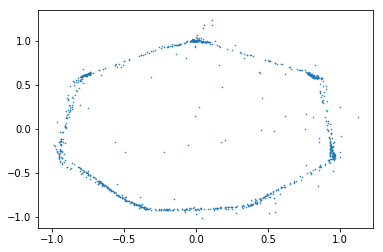}}
\caption[Two numerical solutions]{Left: train dataset of 100 000 points, Right: 100 000 generated samples from the generator after training on 30 epochs}
\label{fig:WGAN_toy}
\end{figure}

\hfill \\
To perform Wasserstein GAN training with gradient penalty, we take the mixture of gaussians described in the previous section. The networks used are also two fully connected networks with two hidden layers. We represent in figure \ref{fig:WGAN_toy} the training data and also the generated samples from the generator after convergence for the first use case. 

We can see here that the generator no longer collapses towards one cluster of the dataset. This example illustrates how well the Wasserstein GAN version focuses on all the data distribution. Nevertheless, convergence is not perfect since there are still generated points outside the data clusters. This is difficult to overcome because of the importance of fine tuning hyperparameters of the models and the choice of gradient penalty constant. 

Moreover, the Discriminator loss is easier to interpret in Wasserstein GAN version since it is proportional to the Wasserstein distance between the true data distribution and the generator distribution. Figure \ref{fig:loss_toyWGAN} is the discriminator loss curve obtained for GAN training: 

\begin{figure}[!htb]
    \centering
    \includegraphics[width=.5\textwidth]{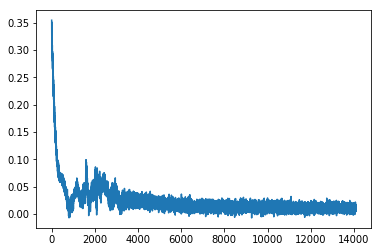}
    \caption{Loss curve of the discriminator in Wasserstein GAN training on multigaussian dataset for 15 epochs}
\label{fig:loss_toyWGAN}
\end{figure}
The loss curve converges towards zero thus the Wasserstein distance between $\mathbb{P}_r$ and $\mathbb{P}_{\theta}$ tends to be close to zero.

\section{Generative Adversarial Networks and anomaly detection} \
Generative Adversarial Networks have been a large subject of study involving several applications. GAN utility can be seen from the generative point of view when the user is interested in producing samples by the generator. Also, GAN have been investigated to perform unsupervised anomaly detection where the goal is to identify "deviant" samples from the original dataset. Despite the several problems encountered during GAN training including mode collapse, many GAN architectures have emerged to perform this kind of anomaly detection. \cite{AnoGAN}, \cite{ADGAN} and \cite{EffiGAN}. \\ The procedure is to train the GAN on normal data and define a score function able to separate normal test data from abnormal data.

\subsection{Anomaly detection based on the discriminator} 
Some anomaly detection methods rely on the discriminator part of the GAN to detect anomalies. Knowing that the discriminator is able to distinguish between real samples drawn from the dataset and fake generated samples given by the generator, the idea behind is to use the resulting network as a classifier for anomalies. \\ This is a very restrictive way of defining what an anomaly can be since the discriminator is only able to detect anomalies that resemble data similar to what the generator was able to produce during training. Moreover, as it was seen in the previous section and proven in \cite{GAN_goodfellow} and \cite{GAN_properties}, the optimal discriminator given a generator has the following expression: 
  \begin{equation}
        D^*_{\theta} \triangleq \frac{\mathbb{P}_r}{\mathbb{P}_r+\mathbb{P}_{\theta}}
    \end{equation}
Suppose we find the best generator parameterization such that $D_{JS}(\mathbb{P}_r ,\mathbb{P}_{\theta}) \approx 0 \ \forall x \in \mathcal{X}$, then the optimal discriminator will always return $1/2 \ \forall x  \in \mathcal{X}$.  \\
Moreover, the author of the original GAN formulation \cite{Goodfellow_cite} states that: 
\begin{displayquote}
For GAN, one thing to keep in mind is that the discriminator is not a generalized detector of weird things. It is trying to tell whether a sample came from the real data or *one specific non-data distribution: the generator*. Because of that, it seems like the discriminator would only be useful for anomaly detection if you think you can make your generator resemble the anomalies you expect to need to detect.
\end{displayquote}
The discriminator seems to only be a tool for the generator to converge towards the real data distribution.

\subsection{Anomaly detection based on the generator} 
Another idea for anomaly detection is to use the generator instead of the discriminator. Since the GAN is trained on only normal data, the generator is only able to produce realistic samples. It is reasonable to assume that if the generator cannot "produce" a certain desired sample, it must be abnormal and considered as an anomaly.
\par
\begin{figure}
    \centering
    \includegraphics[width=.9\textwidth]{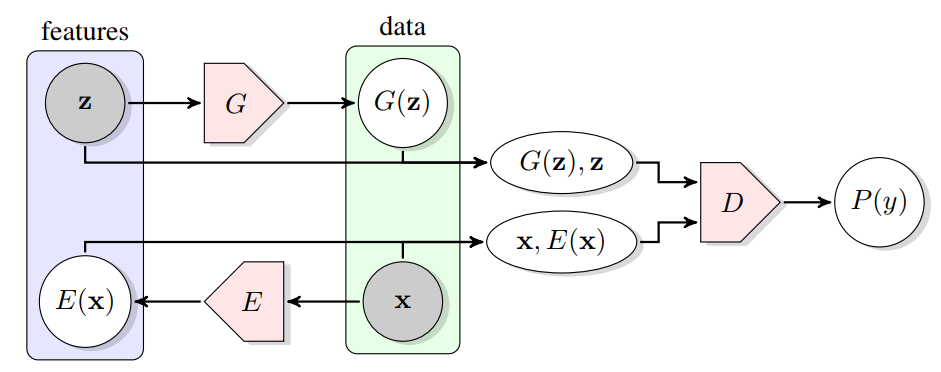}
    \caption{: The structure of Bidirectional Generative Adversarial Networks (BiGAN) \cite{BiGAN}}
\label{fig:BiGAN}
\end{figure}

 Using the generator is less straightforward since the generator does not output an anomaly score function but only fake data samples. Some investigations have been conducted to perform anomaly detection with both the generator and the discriminator. They perform a "reverse mapping" of data to the latent space, the more the reconstruction loss is high the more the queried data can be considered abnormal.     
\begin{itemize}
    \item[$\bullet$]Anomaly GAN (AnoGAN) has been developed to perform anomaly detection on imaging data \cite{AnoGAN}. Given a trained generator $G$ and its corresponding discriminator $D$, AnoGAN defines a reconstruction loss (called residual loss) to be minimized in order to find the closest data sample to the data query: 
    \begin{equation}
        \mathcal{L}_R(z_{\gamma}) = \sum \| x_{query} - G( z_{\gamma}) \|
    \end{equation}
    Under the assumption of a perfect generator G and a perfect mapping to
latent space, for an ideal normal query case, images x and $ G( z_{\gamma})$ are identical. \\
An additional loss was introduced in AnoGAN as a feature matching loss for the discriminator and is supposed to increase performance of anomaly detection: 
\begin{equation}
    \mathcal{L}_D(z_{\gamma}) = \sum \| f_D(x_{query}) - f_D(G( z_{\gamma})) \|
\end{equation}
where $f_D$ is an intermediate layer of the discriminator. This additional loss was inspired by feature matching method described in \cite{improved_training}. \\
The score function is then defined as a combination of both losses: 
\begin{equation}
    \mathcal{L}(z_{\gamma}) = \lambda \mathcal{L}_D(z_{\gamma}) + (1 - \lambda) \mathcal{L}_R(z_{\gamma})  
\end{equation}
The higher the score function the higher the probability that $x$ is abnormal. Nevertheless, in order to perform anomaly detection for each query, AnoGAN needs to solve an optimization problem in the latent space to find the corresponding noise generating the closest sample to the query. This can be an expensive computation that can hardly be avoided considering a standard GAN architecture. \\
\item[$\bullet$]Bidirectional GAN (BiGAN)  \cite{BiGAN} is a framework that can be used for anomaly detection \cite{EffiGAN}. It performs the inverse mapping using an additional neural network (called the encoder $E$) whose role is to learn the inverse function $G^{-1}$ as it is illustrated in Figure \ref{fig:BiGAN}. 
It is also different from the standard GAN setting since the discriminator needs the tuple $(z , G(z))$ or $(E(x) , x)$ as an input and not only $G(z)$ or $x$. The objective function is: 
\begin{equation}
    \min_{G,E} \ \max_{D} V (D, E, G)
\end{equation}
where 
\begin{multline}
    V(D,E,G) =  \mathbb{E}_{x \sim p(X)}[   \mathbb{E}_{z \sim p_E(.|x)}  [\log D(x, z)] ] +\\ \mathbb{E}_{z \sim p(Z)}[   \mathbb{E}_{z \sim p_G(.|z)}  [1-\log D(x, z)] ]
\end{multline}
BiGAN also defines a reconstructed loss function:
\begin{equation}
    L_G(x)=\| x - E(G(x))\|
\end{equation}
and discriminator loss based on feature matching:
\begin{equation}
    L_D = \| f_D(x,E(x)) - f_D(G(E(x)),E(x))\|
\end{equation}
The score function is defined as : \\
\begin{equation}
    A(x) = \alpha L_G(x) + (1-\alpha)L_D(x)
\end{equation}
Again, the higher the score function the higher the probability of $x$ being abnormal
\end{itemize}

\section{Contribution}
Based on this literature review, we chose to perform anomaly detection using a Wasserstein Generative Adversarial Network. The main reason is that Wasserstein GAN does not collapse contrarily to the classical GAN which needs to be heavily tuned in order to avoid this problem. Mode collapse can be blocking if we need to perform anomaly detection: if a subset of our data distribution is not learned by the generator, then all samples that are similar to this subset might end up classified as abnormal. Another added value of the wasserstein GAN version compared to a standard GAN is the possibility of using the loss function of the discriminator to evaluate convergence since it is an approximation of the Wasserstein distance between $\mathbb{P}_r$ and $\mathbb{P}_{\theta}$. With standard GAN, it is more difficult to interpret the loss curve.
\\ \par We investigate abnormal data on the multi-gaussian distribution from \ref{ssec:UC1} and also in higher dimensional space from \ref{ssec:UC2} following the same procedure as in \cite{EffiGAN} and \cite{VAE} . We split the MNIST dataset in $80\%$ of normal training data and $20\%$ plus all abnormal data for testing. We define normality as all digits except the one we considered abnormal. Each class is considered abnormal and 10 models with corresponding scores were defined as in \cite{EffiGAN} and \cite{VAE}. 
\par To evaluate our model, we compute the Precision-Recall curves and also the Area Under Precision Recall Curve (AUPRC) and compare our results to standard anomaly detection methods based on GAN and Variational-Auto-Encoders.

\subsection{Anomaly detection with the Critic}
We first want to perform anomaly detection with the critic (discriminator) of a W-GAN. First, we train a Wasserstein GAN on use case 1. The networks used are 2 fully connected neural networks with a 2-128-128-2 architecture for the generator and a 2-128-128-1 architecture for the discriminator (also called the critic). The latent space (generator input) takes samples from a standard normal distribution $\mathcal{X}$ in 2 dimensions. 
\begingroup
\begin{figure}[!htbp]
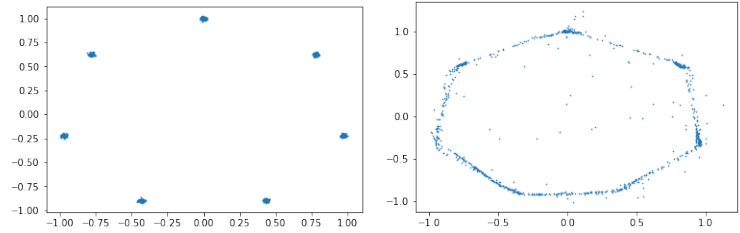

\centering
  \subfigure{\includegraphics[width=.4\linewidth]{multigaussian.png}}
  \subfigure{\includegraphics[width=.4\linewidth]{multigau_gene.png}}
\caption[Two numerical solutions]{Left: train dataset of 100 000 points, Right: 100 000 generated samples from the generator after training on 30 epochs}
\label{fig:WGAN_UC2}
\end{figure}
\endgroup

\begin{figure}[!htbp]
\centering
\subfigure{\includegraphics[height=4cm,width=.4\linewidth]{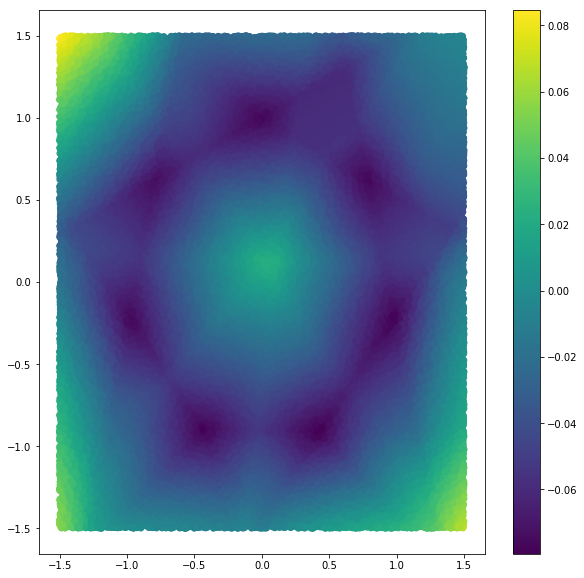}}
\subfigure{\includegraphics[height=4cm,width=.4\linewidth]{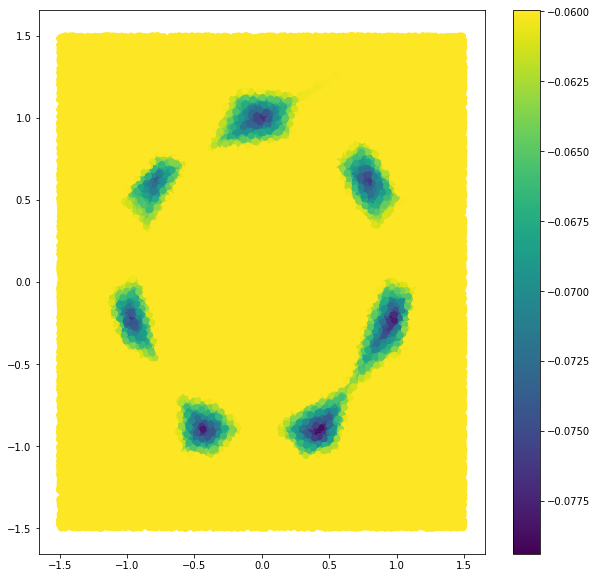}}
\caption[Two numerical solutions]{Left: Ouput of the critic on $\mathcal{X}$. Right: output on $\mathcal{X}$ clipped around the confidence interval given by the range of the critic output on the training dataset}
\label{fig:critic_output}
\end{figure}

 It is highly difficult for a standard GAN to converge on this dataset since mode collapse doesn't allow the generator to catch all the clusters of the distribution \cite{Unrolled_GAN} . The generator of the W-GAN learns the distribution but fails on providing realistic samples because there are still points generated outside the given distribution as it can be seen in Figure \ref{fig:WGAN_UC2}. This is a limitation of the W-GAN algorithm that might be alleviated with better parameteization. Nevertheless, convergence is sufficient if we want to perform anomaly detection because we can still identify points outside the resulting distribution. We plot the output of the critic in the the space $\mathcal{X}$ in order to visualize its utility for anomaly detection. Results as displayed in Figure \ref{fig:critic_output}.
 
On this dataset, we observe that the critic might be useful to detect anomalies. We define a confidence interval on the output of the critic as being the interval between the minimum and the maximum value of the output on the training dataset. Every point outside the confidence interval might be considered as an anomaly. 
\\ 
\par Unfortunately, this technique doesn't work very well on use case 2.  We train a WGAN with convolutional generator and critic until convergence on digits from 1 to 9 to make the 0 resemble an abnormal data, and observe the output of the critic on the test dataset:  
\begin{figure}[!htbp]
    \centering
    \includegraphics[width=0.7\textwidth]{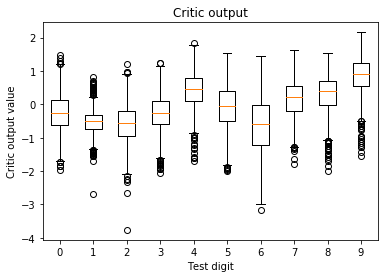}
    \caption{Output of the critic on MNIST test dataset}
\label{fig:my_failure_uc2}
\end{figure}\\
The critic fails to separate the 0 from the other digits as it can be noticed in Figure \ref{fig:my_failure_uc2}. We think this might be related to the curse of dimension since an MNIST digit lies on a 784-dimensional space whereas a data from the mixture of gaussians is only 2-dimensional. Moreover, the usage of the critic alone as an anomaly detector is not very clear from a theoretical point of view. We decide to investigate the utility of the generator for anomaly detection, which is more intuitive.  
\subsection{Reversing the generator}

\subsubsection{Objective}\hfill\\
In order to inverse the generator and find if it is capable of producing samples resembling the query data, we can think of solving an optimization problem as in \cite{InverseG}. If $x$ is the query sample that we want to inverse, the problem formalizes as follows. 
\begin{equation}
     z^* = \underset{z}{\min}-\mathbb{E}_x[\log(G(z))]
\end{equation}

Nevertheless, we are not only interested on finding if there is a possible input for the generator that will produce a sample resembling $x$, we also want to make sure that this input can be produced from sampling from the latent space distribution $\mathcal{N}(0,1)$. In order to achieve this, we penalize the objective function with a term $\log P(\mathbf{z}) = 1/d \sum_i \log(P(z_i))$ if $P$ is the normal density function \cite{InverseG}. \\
The new loss function to be minimized is:
\begin{equation}
    L_x(z) = \mathbb{E}_x[\log(G(z))] - \lambda \log P(\mathbf{z})
    \label{eqn:reverse_gen}
\end{equation}
By fine tuning the penalization constant $\lambda$ and by performing gradient descent on equation \ref{eqn:reverse_gen}, we achieve convergence that helps insuring that the optimal $z^*$ comes from probable regions of the latent space. 
\subsubsection{Experiment}
\begin{figure}[!htb]
    \centering
    \includegraphics[width=.7\textwidth]{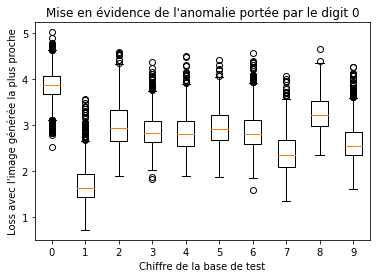}
    \caption{Reversed test data losses of a W-GAN trained on digits from 1 to 9}
\label{fig:ReverseGAN}
\end{figure}

\hfill\\
We train the W-GAN on MNIST digits from 1 to 9 and perform the inverse mapping. The boxplot in Figure \ref{fig:ReverseGAN} computes the mean squared error between each test sample in the database in its reconstitution. More than half of the zero digits are well separated from the other test samples. We can already notice that our model captures the fact that the reconstitution of the digit $0$ is not achieved as well as for all other test samples and as a result can be considered abnormal. 
\\
\par Unfortunately, an expensive trade off between number of iterations for gradient descent and results precision has to be made which can considerably impact the inference time for each query $x$.

\subsection{Encoder-decoder GAN}
\subsubsection{Objective} \hfill\\
As in AnoGAN, performing anomaly detection by solving an optimization problem for each query data can be expensive and/or inefficient, especially if one needs to perform this kind of task in real-time applications where inference is crucial. The simple idea is to use an additional network to learn the inverse mapping, increase training time and reduce inference time. But contrarily to what has been investigated in \cite{BiGAN} and \cite{EffiGAN}, we train the encoder after finishing the training of the GAN and not simultaneously whereas \cite{BiGAN} has to do the training of E and G at the same time because it is specified by the objective function. \\
We define the training loss function of our encoder as : 
\begin{equation}
    \min_{\alpha} \sum_i ( G_{\theta}(E_{\alpha}(x_i)) - x_i )^2 \ \forall \textbf{x} \in \mathcal{X}
\end{equation}
Where $\theta$ are the trained generator weights and $\alpha$ the encoder weights to be optimized. \\
\par Moreover, in order to ensure that the encoder encodes data well in the latent space, we force the input of the latent space (ie. the encoder output) to have a 0 mean and 1 variance by adding a Batch Normalization layer. This helps the encoder to only look for the reconstitution in the "normal" areas of the latent space which correspond to the normal distribution $\mathcal{N}(0,1)$.

\subsubsection{Experiment on use case 2}\hfill\\
As in \cite{EffiGAN}, we created 10 different models from MNIST by successively making each digit class
an anomaly and treating the remaining 9 digits as normal examples. 
\par First, we train a standard Wasserstein GAN without the encoder part until convergence, after converging we train an Autoencoder model were a neural network (the encoder) is stacked with the trained generator.  
\par
\begin{figure}[!htbp]
\centering
\begin{minipage}{.35\linewidth}
  \subfigure{\includegraphics[width=1\linewidth]{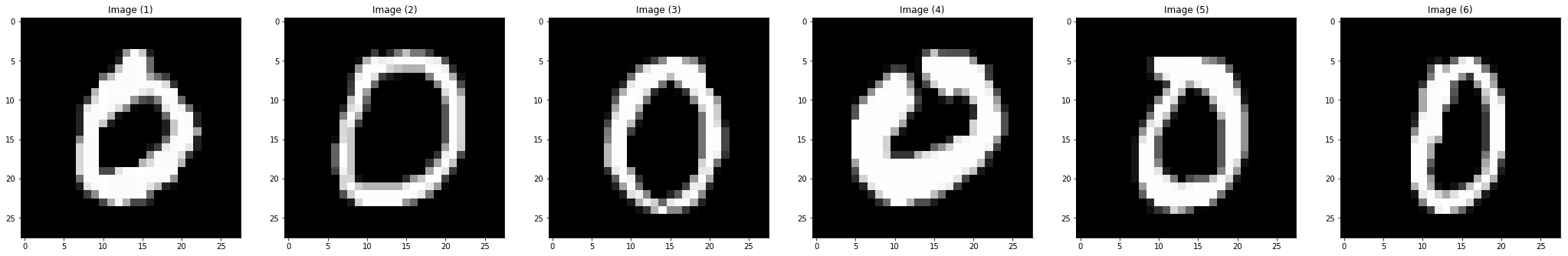}}
  \subfigure{\includegraphics[width=1\linewidth]{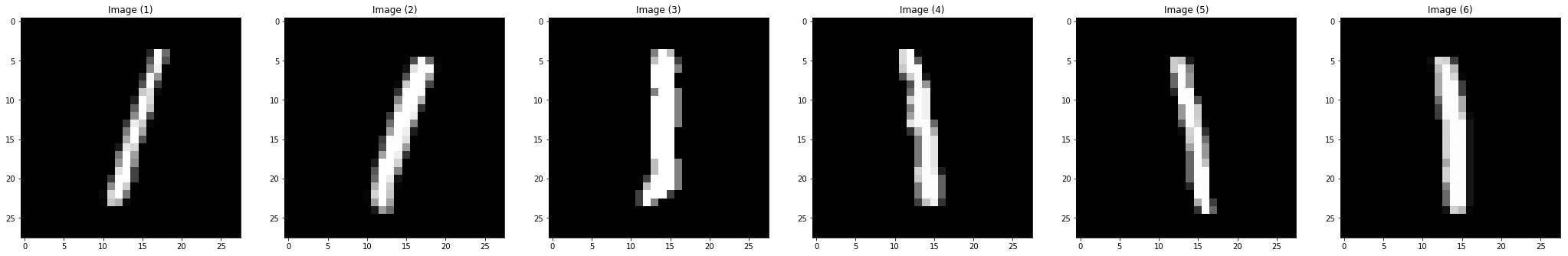} }
  \subfigure{\includegraphics[width=1\linewidth]{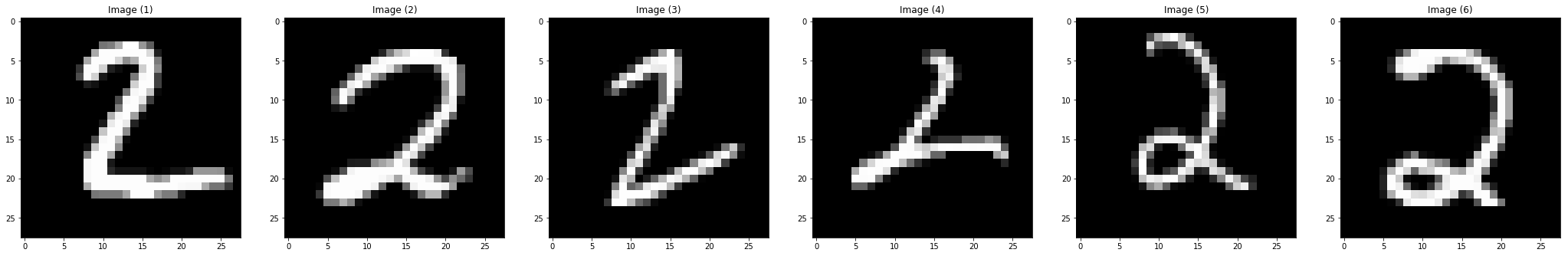}}
  \subfigure{\includegraphics[width=1\linewidth]{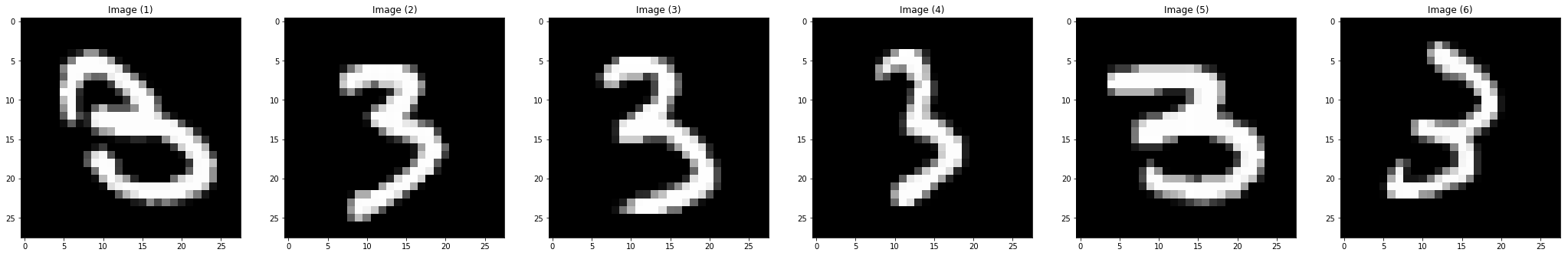}}
  \subfigure{\includegraphics[width=1\linewidth]{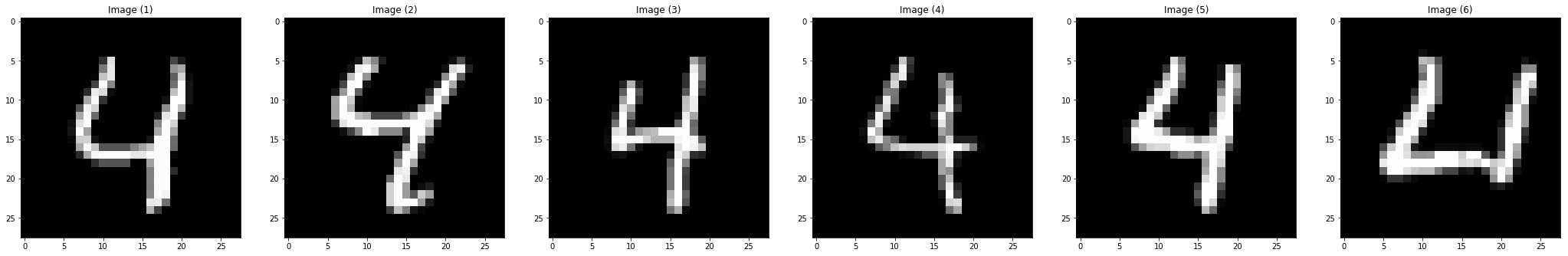}}
  \subfigure{\includegraphics[width=1\linewidth]{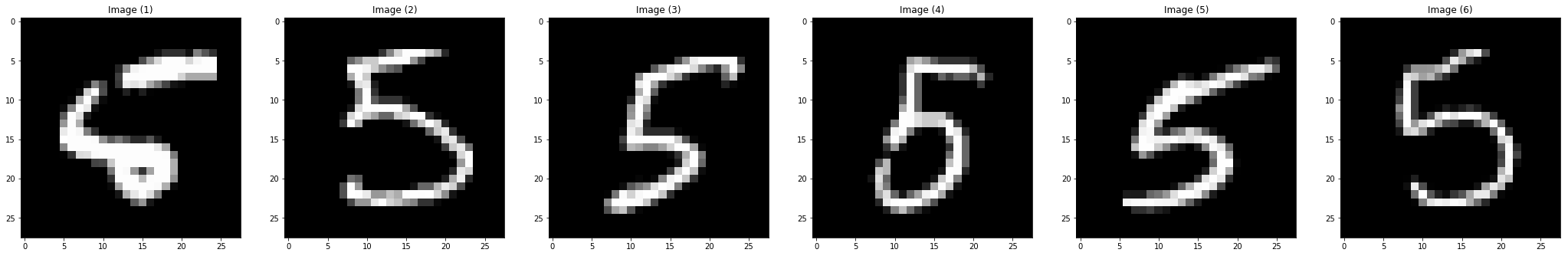} }
  \subfigure{\includegraphics[width=1\linewidth]{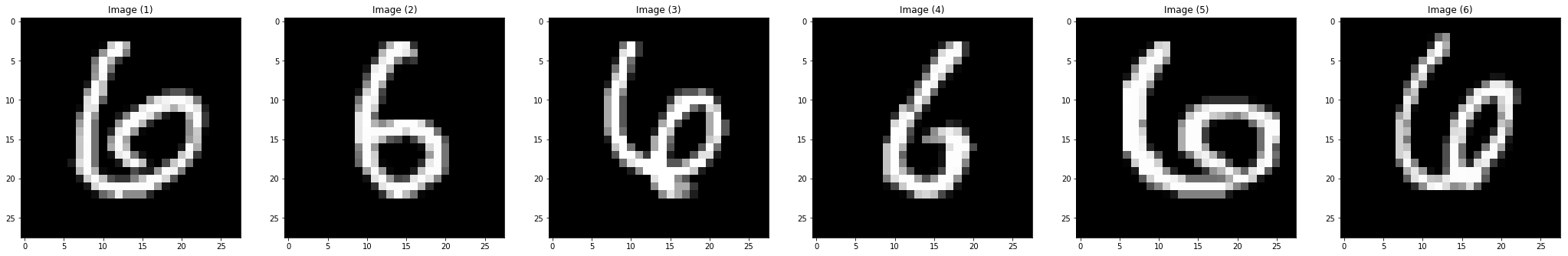}}
  \subfigure{\includegraphics[width=1\linewidth]{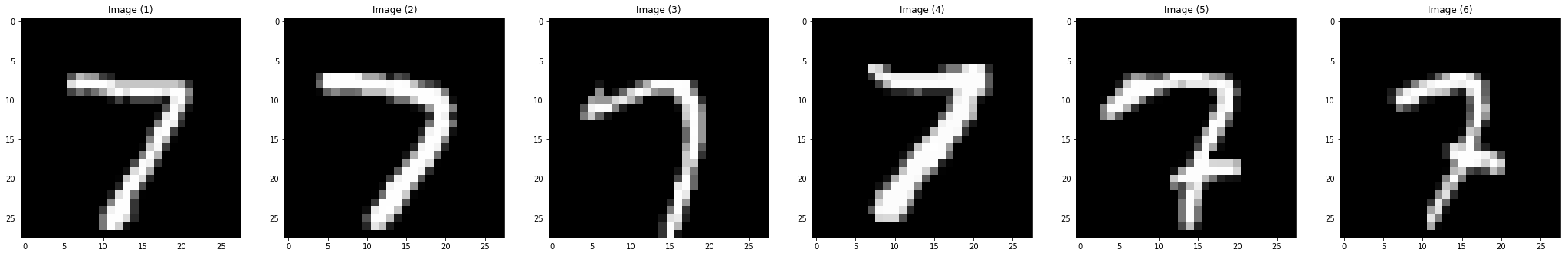}}
  \subfigure{\includegraphics[width=1\linewidth]{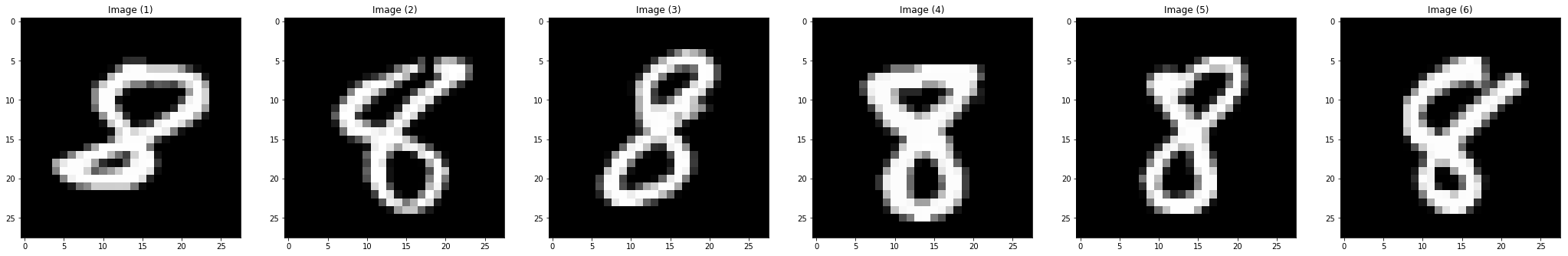}}
  \subfigure{\includegraphics[width=1\linewidth]{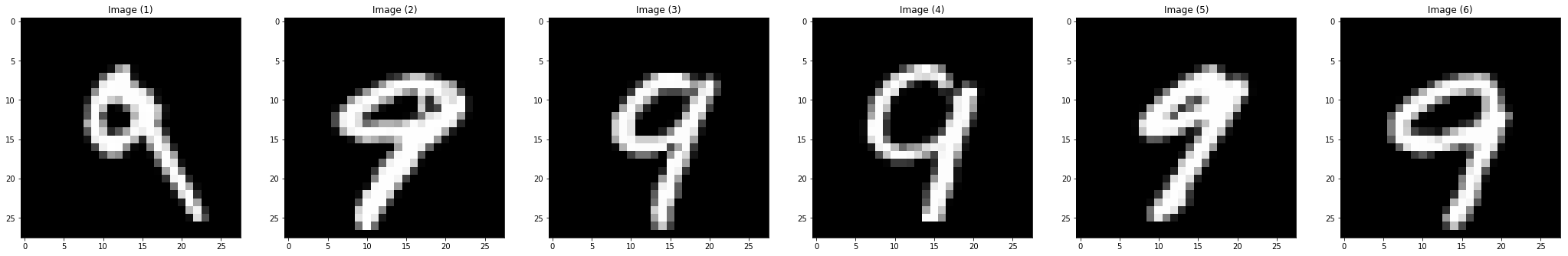}}
\end{minipage}
\begin{minipage}{0.35\textwidth}
  \subfigure{\includegraphics[width=1\linewidth]{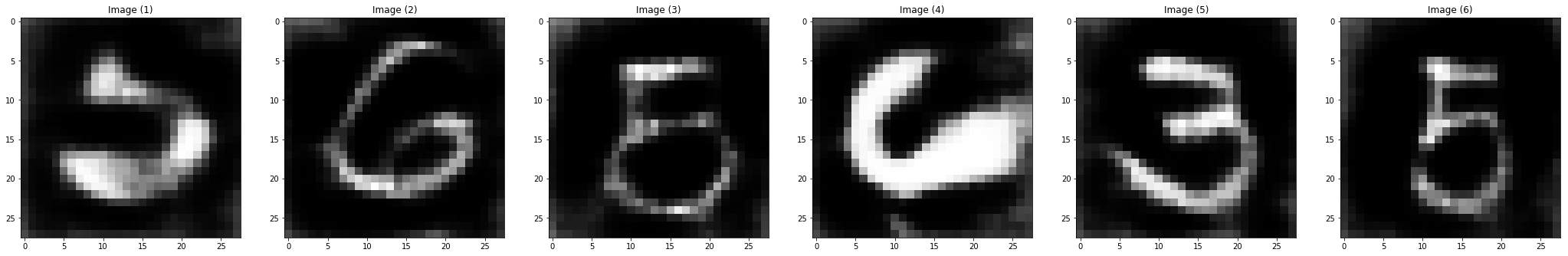}}
  \subfigure{\includegraphics[width=1\linewidth]{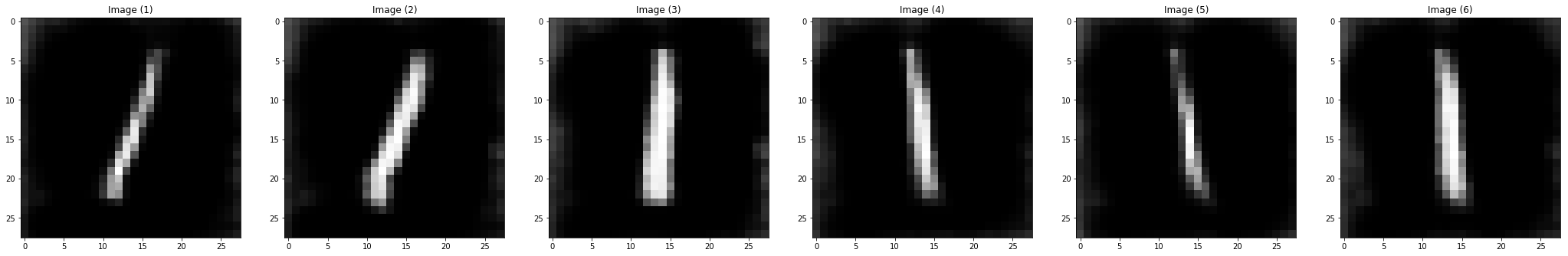}}
  \subfigure{\includegraphics[width=1\linewidth]{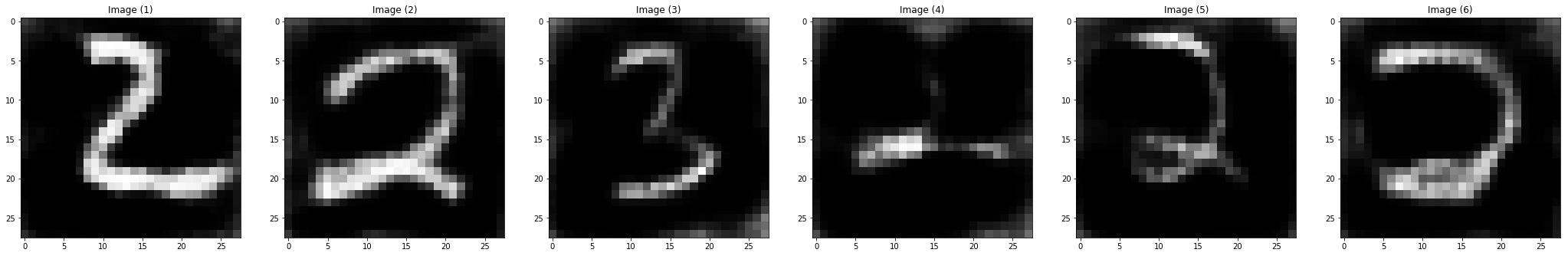}}
  \subfigure{\includegraphics[width=1\linewidth]{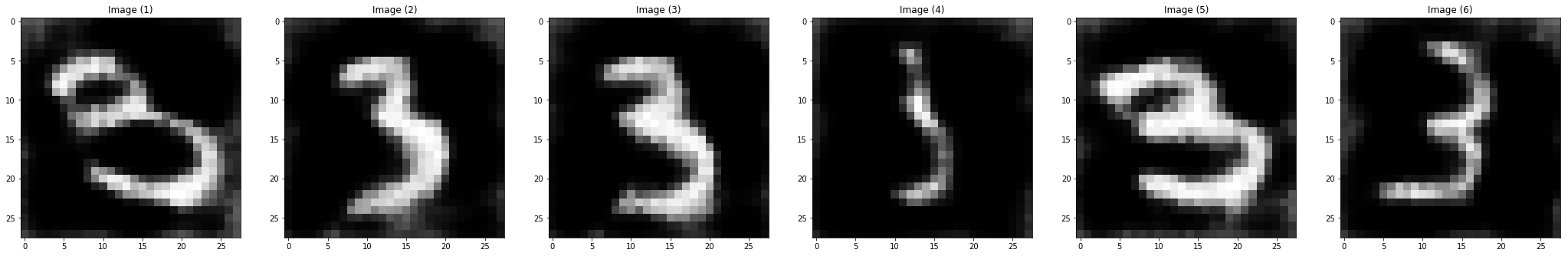}}
  \subfigure{\includegraphics[width=1\linewidth]{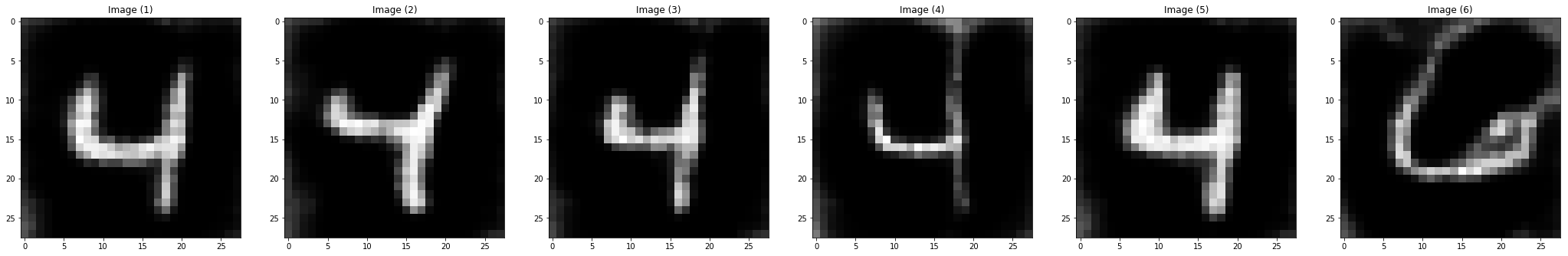}}
  \subfigure{\includegraphics[width=1\linewidth]{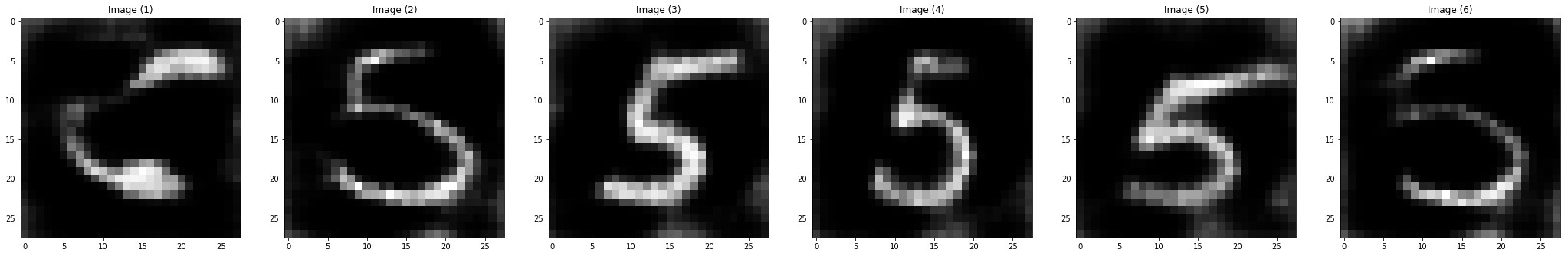} }
  \subfigure{\includegraphics[width=1\linewidth]{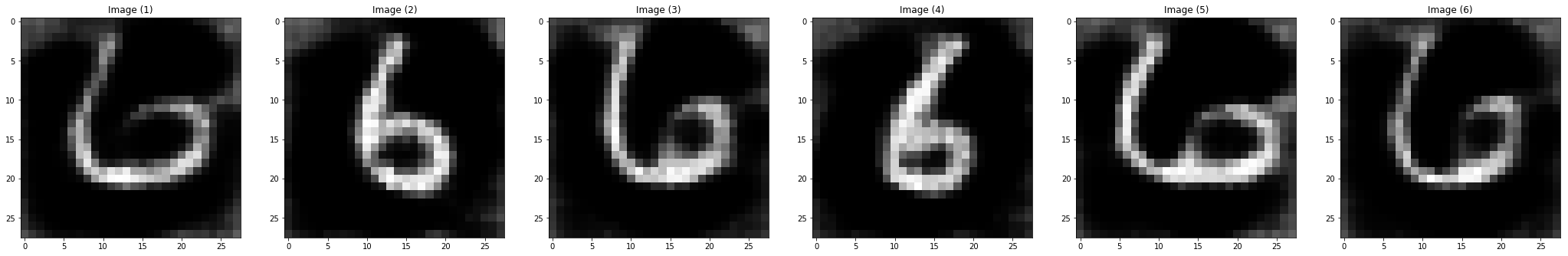}}
  \subfigure{\includegraphics[width=1\linewidth]{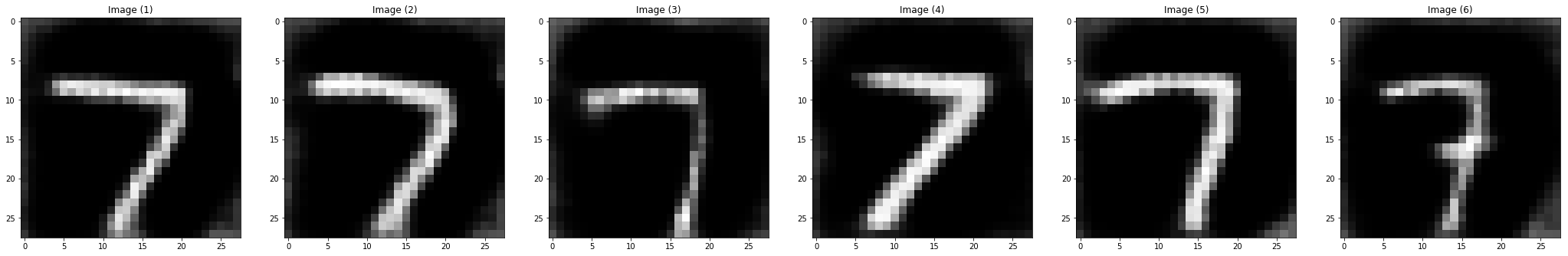}}
  \subfigure{\includegraphics[width=1\linewidth]{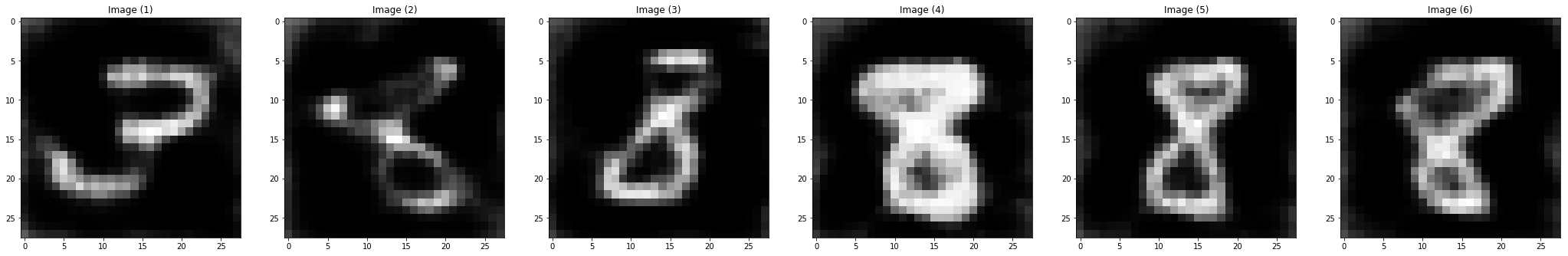}}
  \subfigure{\includegraphics[width=1\linewidth]{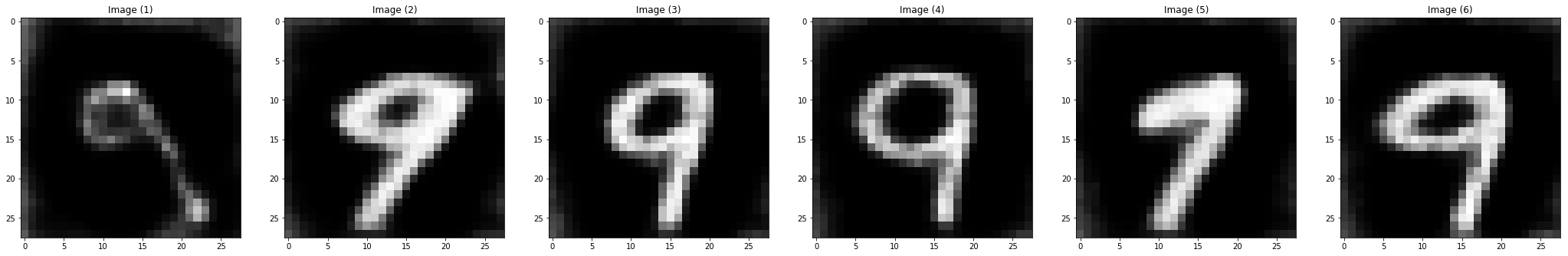}}
\end{minipage}
\caption[Two numerical solutions]{Wasserstein GAN with Encoder model trained on digits from 1 to 9. First column: test data, second column: their reconstitution }
\label{fig:0rec}
\end{figure}

\begin{figure}[!htbp]
\centering

\begin{minipage}[b]{0.35\textwidth}
  \subfigure{\includegraphics[width=1\linewidth]{0.png}}
  \subfigure{\includegraphics[width=1\linewidth]{1.png} }
  \subfigure{\includegraphics[width=1\linewidth]{2.png}}
  \subfigure{\includegraphics[width=1\linewidth]{3.png}}
  \subfigure{\includegraphics[width=1\linewidth]{4.png}}
  \subfigure{\includegraphics[width=1\linewidth]{5.png} }
  \subfigure{\includegraphics[width=1\linewidth]{6.png}}
  \subfigure{\includegraphics[width=1\linewidth]{7.png}}
  \subfigure{\includegraphics[width=1\linewidth]{8.png}}
  \subfigure{\includegraphics[width=1\linewidth]{9.png}}
\end{minipage} 
\begin{minipage}[b]{0.35\textwidth}
  \subfigure{\includegraphics[width=1\linewidth]{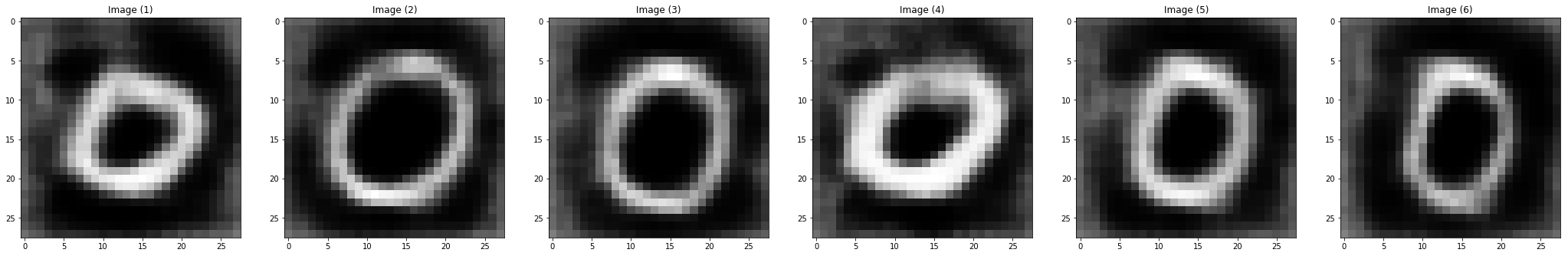}}
  \subfigure{\includegraphics[width=1\linewidth]{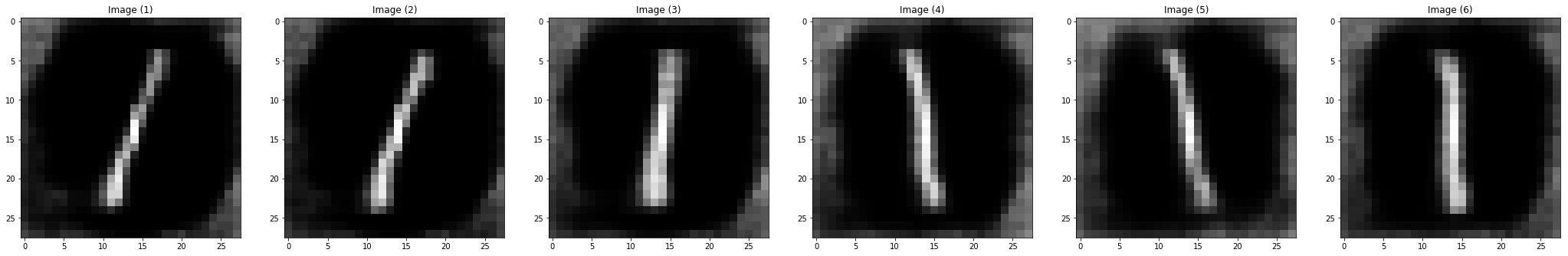}}
  \subfigure{\includegraphics[width=1\linewidth]{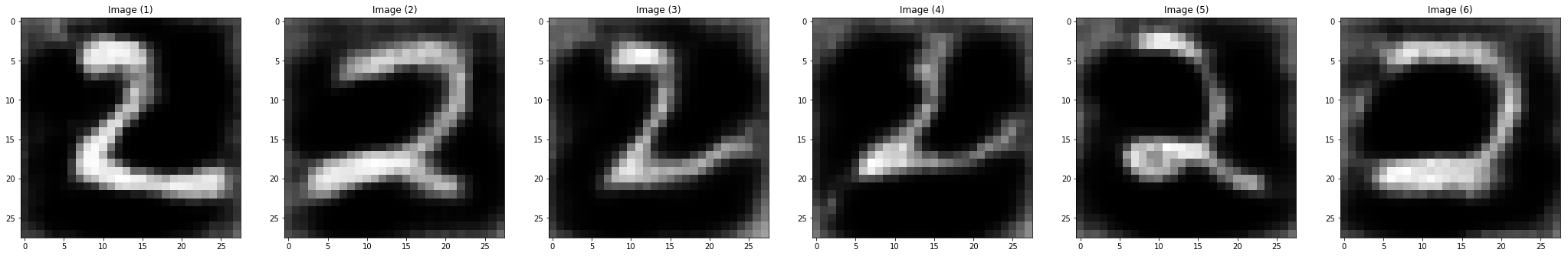}}
  \subfigure{\includegraphics[width=1\linewidth]{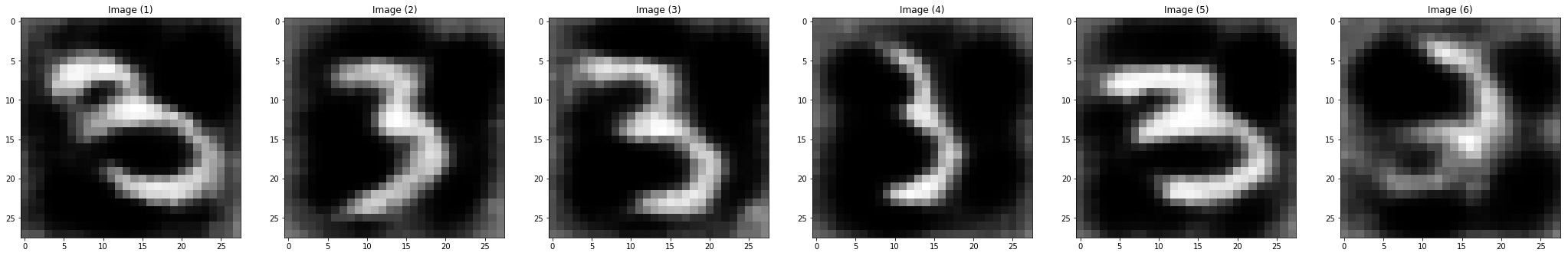}}
  \subfigure{\includegraphics[width=1\linewidth]{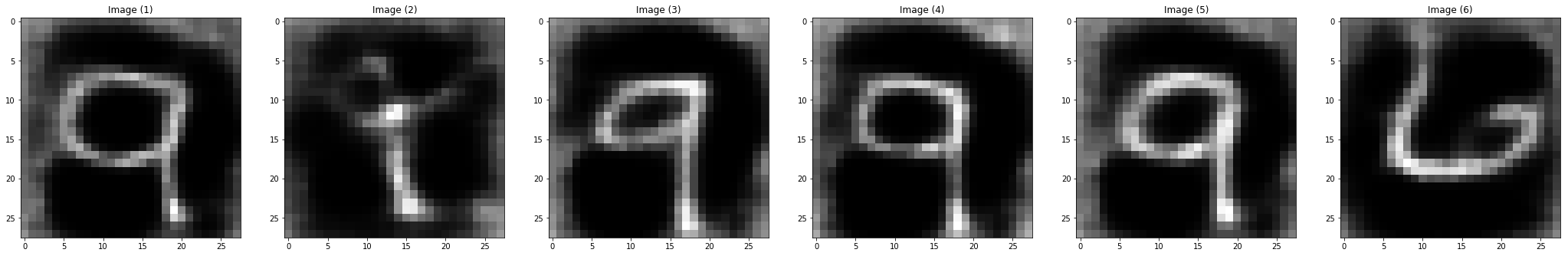}}
  \subfigure{\includegraphics[width=1\linewidth]{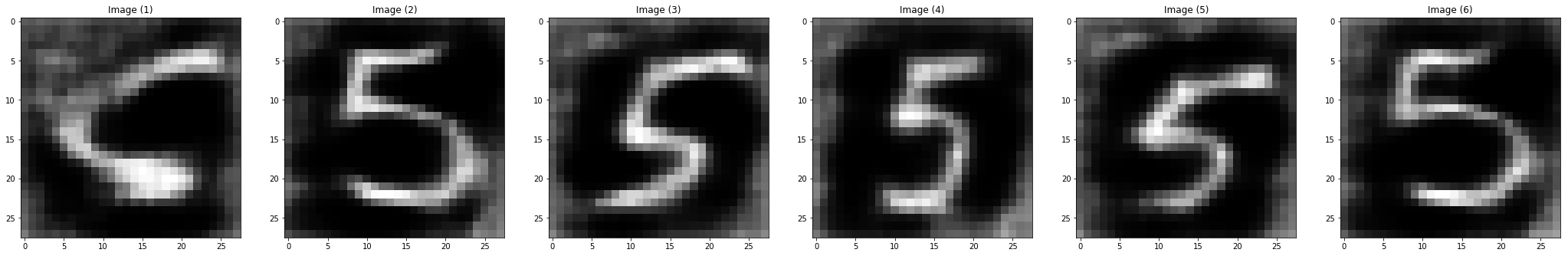} }
  \subfigure{\includegraphics[width=1\linewidth]{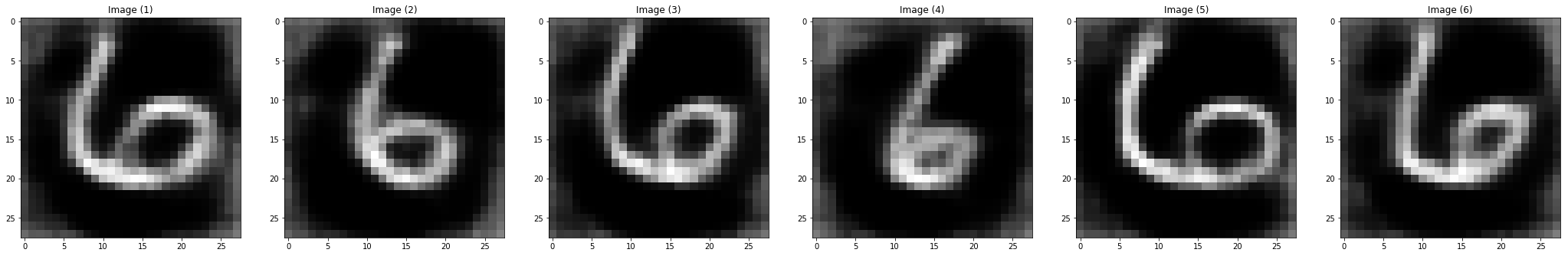}}
  \subfigure{\includegraphics[width=1\linewidth]{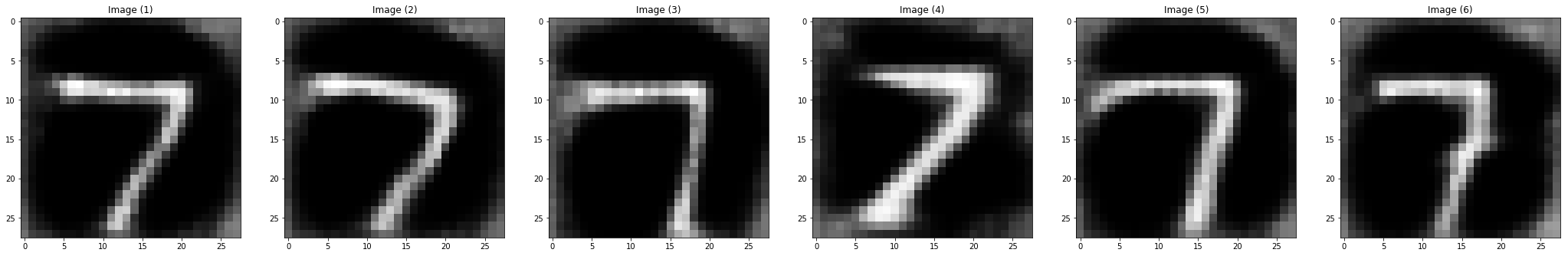}}
  \subfigure{\includegraphics[width=1\linewidth]{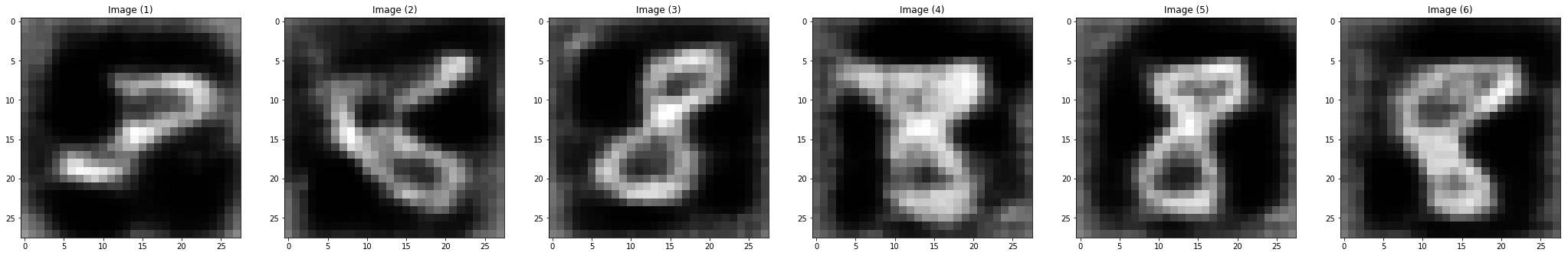}}
  \subfigure{\includegraphics[width=1\linewidth]{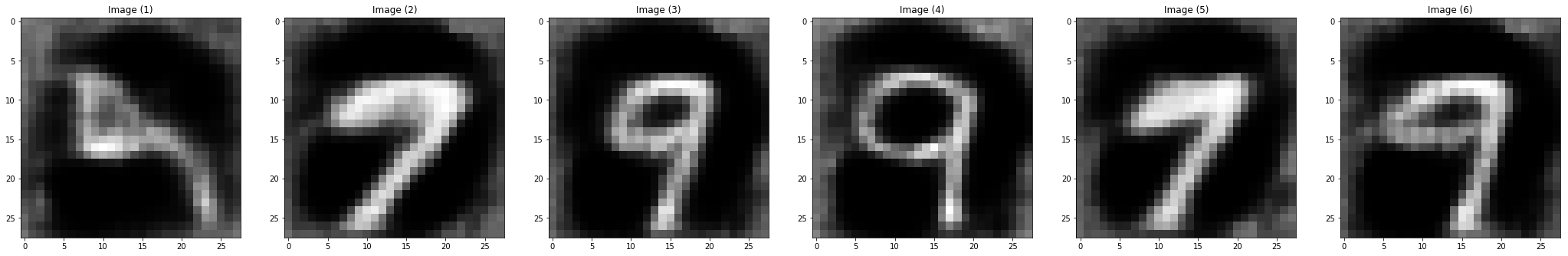}}
\end{minipage} 
\caption[Two numerical solutions]{Wasserstein GAN with Encoder model trained on all digits except the digit 4. First column: test data, second column: their reconstitution}
\label{fig:quatrerec}
\end{figure}

 Figures \ref{fig:0rec} and \ref{fig:quatrerec} show the reconstituted digits from the test dataset. Figure \ref{fig:0rec} shows that the Autoencoder-WGAN trained on all digits except the 0 fails in reconstituting this digit, which is the desired property. Figure \ref{fig:quatrerec} shows that a model that wasn't trained on digit 4 is guessing 9 instead which is the kind of behaviour we are looking for.

\par Finally, We compute the mean squared error between each test sample and the closest generated sample from the Autoencoder. The results obtained for each class considered as being abnormal are represented in Figure \ref{fig:mse_mnist} of Appendix A.

The Wasserstein GAN with encoder loss separates well the normal samples from the abnormal ones. This can be seen in the boxplot where the right box is higher than the left one for each digit considered abnormal. To gain more insight, we define moving thresholds for the loss that are able to separate normality from abnormality. Precision Recall scores are then calculated for each threshold and the area under precision recall is computed for abnormal class. \\

\begin{table}[]
    \centering
    \begin{tabular}{|c|c|c|c|}
         \hline
  Abnormal digit & AUPRC VAE & AUPRC BiGAN & AUPRC our model \\
  \hline
  0 & 51.7 \% & 80 \% & \textbf{97} \% \\
  1 & 6.3 \% & 30 \% & \textbf{51} \% \\
  2 & 64.4 \% & 70 \% & \textbf{89} \% \\
  3 & 25.1 \% & 55 \% & \textbf{78} \% \\
  4 & 33.7 \% & 50 \% & \textbf{83} \% \\
  5 & 32.5 \% & 55 \% & \textbf{72} \% \\
  6 & 43.2 \% & 63 \% & \textbf{87} \% \\
  7 & 14.8 \% & 40 \% & \textbf{57} \% \\
  8 & 49.9 \% & 57 \% & \textbf{90} \% \\
  9 & 10.4 \% & 38 \% & \textbf{70} \% \\
 
  \hline
  
    \end{tabular}
    \caption{Area Under Precision Recall Curves (AUPRC) for each digit considered an anomaly}
    \label{tab:AnoAUPRC}
\end{table}

\par To test our model, the AUPRC is compared to the ones obtained with Variational Auto-Encoders and BiGAN. We can see from the table \ref{tab:AnoAUPRC} above that our model outperforms most of usual GAN algorithms for anomaly detection. Results for VAE model have been taken from \cite{VAE}, and results of BiGAN \cite{EffiGAN} have been taken from the curve in appendix A. In our opinion, this is due to the fact that the Wasserstein GAN captures the data distribution well which is the most important property that the Generator has to satisfy if its goal is to capture the "normality" in the training data set. 

\section{Towards anomaly detection on use case 3}

\par The next step in our study is to investigate anomalies on multivariate time series data. The UCI dataset presented in the second section constitutes our use case and we look forward to apply this method to detect anomalies. One can think of different network architectures to perform anomaly detection on time series: Convolutional neural networks, Recurrent networks, Multilayer perceptrons... 
Unfortunately, the gradient penalty of the Improved Wasserstein GAN doesn't allow the exploitation of recurrent neural networks since they throw errors during the calculation of second order derivatives using most deep learning libraries such as keras and tensorflow. We decided to investigate the CNNs to perform the task. \\
Each time series have been translated into an image using a distance plot matrix (similar to the recurrence plot matrix) defined as in \cite{TS_image}:\\
\begin{equation}
    R_{i,j} = \| \mathbf{s_i} - \mathbf{s_j} \| 
\end{equation}
Where $\mathbf{s_i}$ is the considered time series evaluated at instant $i$. Figure \ref{fig:im_ts} represents a time series with its corresponding recurrence image: 

\begin{figure}[!htbp]
\centering
  \subfigure{\includegraphics[height=4cm,width=.45\linewidth]{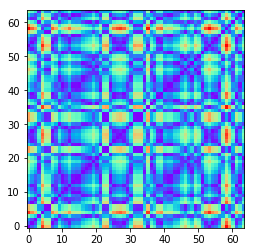}}
  \subfigure{\includegraphics[height=4cm,width=.45\linewidth]{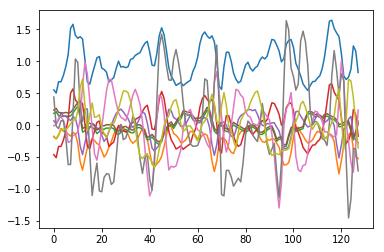}}
\caption[Two numerical solutions]{ Left: Recurrence plot for blue curve without threshold (distance matrix); Right: time series data of walking individual}
\label{fig:im_ts}
\end{figure}

\begin{figure}[!htb]
    \centering
    \includegraphics[width=.7\textwidth]{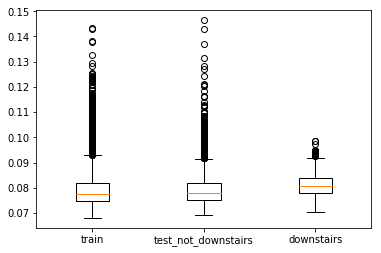}
    \caption{Boxplot of reconstitution errors}
\label{fig:boxts}
\end{figure}

Then we treat the multivariate time series as a multi-channel image with 9 channels and apply the Wasserstein GAN with encoder procedure. We consider the class "walking downstairs" as being abnormal. As for the second use case, we compute the mean squared error between each query sample from the test dataset and its reconstitution. We obtain the boxplots of Figure \ref{fig:boxts}. 

\begin{figure}[!htbp]
\centering
  \subfigure{\includegraphics[height=4cm,width=.45\linewidth]{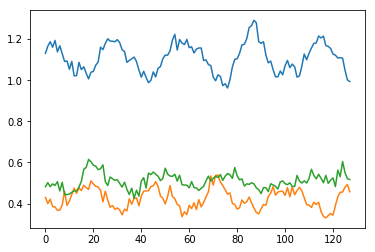}}
  \subfigure{\includegraphics[height=4cm,width=.45\linewidth]{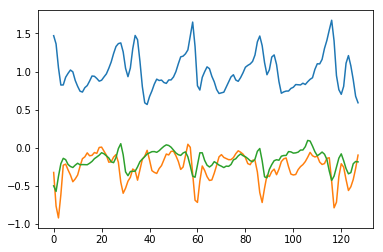}}
\caption[Two numerical solutions]{ Left: Corresponding reconstitution from the model; Right: time series data of walking individual}
\label{fig:recoFCN}
\end{figure}

The reconstitution is not high enough for the abnormal class to clearly separate from the normal one. We cannot deduce from this result that a specific query data is normal or not. \\
Another approach consists in considering only fully connected neural networks as models. We train the Wasserstein GAN on the total acceleration raw data which consists in 3 time series. Then we train the encoder and watch the reconstitution of training data. The quality of the generated samples hardly matches the training dataset.  Figure \ref{fig:recoFCN} shows a walking activity acquisition and its corresponding reconstitution.\\

\begin{figure}[!htb]
    \centering
    \includegraphics[width=.7\textwidth]{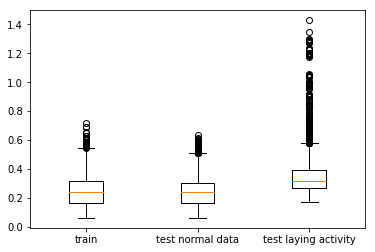}
    \caption{Boxplot of reconstitution errors}
\label{fig:anofcn}
\end{figure}

Nevertheless, we compare the mean squared error between the normal test samples and their reconstitution, and the abnormal test data (laying activity) and their reconstitution. Figure \ref{fig:anofcn} shows these errors on a boxplot. \\

This boxplot allows the computation of the precision recall curves and the AUPRC. The model is not perfect but manages to achieve an AUPRC score of $70 \%$ in the detection of the abnormal class. 

A future improvement consists in considering CNN for both the generator and discriminator in order to detect anomalies from raw time series data. 1-D convolutions are needed and will be investigated to produce good visual representations of time series samples.A more thorough study of the impact of the architecture should also be done. \\

\section{Conclusion and way forward}
In this paper, we investigate the added value of generative adversarial networks in the task of anomaly detection. We demonstrated the ability of Wasserstein Generative Adversarial Networks to detect anomalies in a high dimensional space. The usage of the discriminator part of the GAN has shown weak properties to efficiently detect abnormalities, whereas the generator seems to be useful in this task. Reversing the generator is then necessary and can be achieved by mapping each query data to the latent space with an optimization scheme or by defining an encoding network to perform this task since inference cost is important.\\

\par In this work, we also search for theoretical properties of GAN able to perform anomaly detection. Despite some recent article providing insights, there are many questions that remain unanswered. There are of course interrogations related to deep learning architectures where it is difficult to understand which network performs well with which dataset. Moreover, GAN rely on the definition of a latent space where the dimension is only defined empirically. Also, the choice of the distribution of this latent space is still an open question. In our experiments we sample from $\mathcal{N}(0,1)$ to feed the latent space. When training the encoder with the generator we noticed that batch normalization before the input layer of the generator helps the convergence of the encoder. This is questionable since batch normalization only transform the mean to 0 and the variance to 1, this procedure gives relatively good results but a formal projection into the latent space may improve convergence. \\

\section*{Acknowledgements}
I would like to express my gratitude to my Airbus supervisor Jayant Sen Gupta for his dedicated support, useful remarks and for providing me with the opportunity to take part in internal workshops. I would like to thank the XRD team of Airbus in the CRT department from ADvISED project which focuses on anomaly detection with Vincent Feuillard and Olivier Regnier Coudert but also all the other members of the data science research team. Special thanks to the 0-AOG plateau of Airbus and its well-known Techno Coffee organized by Fabrice Jimenez. Also, I am grateful to my Supaero data science teacher Emmanuel Rachelson for his classes. Last but not least, special thanks to the other interns and friends who supported me during this project, and to my family for their unconditional backing. 

%
%
%
%

\newpage
\bibliographystyle{abbrv}
\bibliography{refs}

\newpage
\appendix
\section{Reconstitution error of MNIST test digits}
\begin{figure}[H]

\centering  

  \subfigure{\includegraphics[height=3.2cm ,width=.45\linewidth]{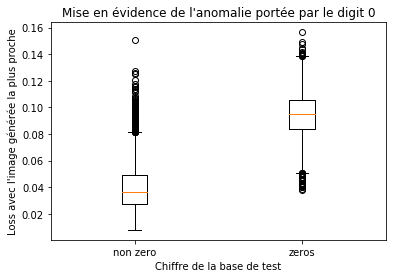}}
  \subfigure{\includegraphics[height=3.2cm ,width=.45\linewidth]{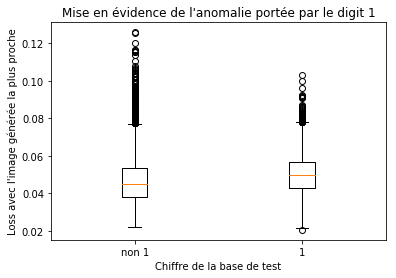}}
  \subfigure{\includegraphics[height=3.2cm ,width=.45\linewidth]{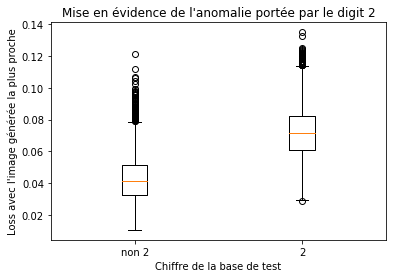}}
  \subfigure{\includegraphics[height=3.2cm ,width=.45\linewidth]{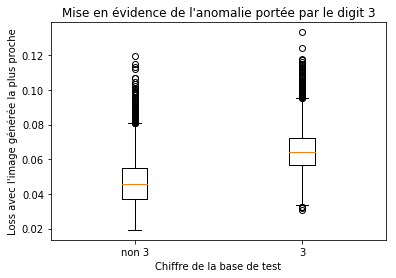}}
  \subfigure{\includegraphics[height=3.2cm ,width=.45\linewidth]{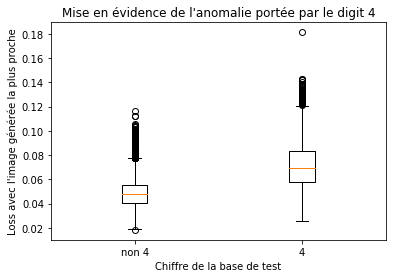}}
  \subfigure{\includegraphics[height=3.2cm ,width=.45\linewidth]{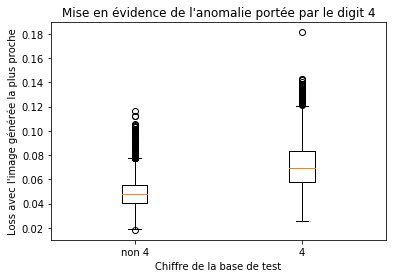}}
  \subfigure{\includegraphics[height=3.2cm ,width=.45\linewidth]{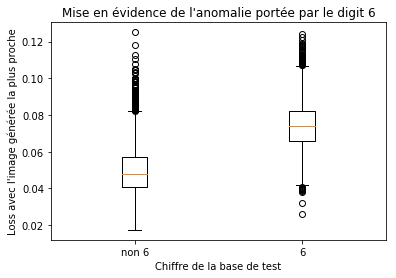}}
  \subfigure{\includegraphics[height=3.2cm ,width=.45\linewidth]{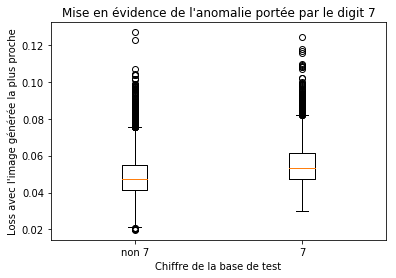}}
  \subfigure{\includegraphics[height=3.2cm ,width=.45\linewidth]{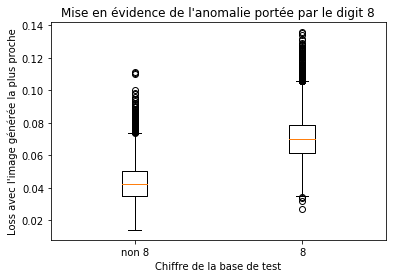}}
  \subfigure{\includegraphics[height=3.2cm ,width=.45\linewidth]{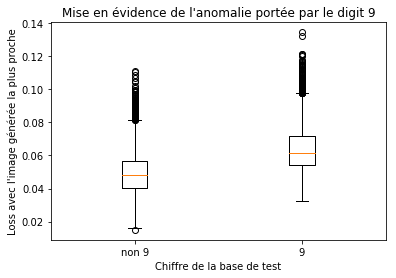}}

\caption{Reconstitution error of MNIST test digits. From top left to bottom right: Reconstitution error of normal digit (left box) against reconstitution of abnormal test digits (right box) from 0 to 9}
\label{fig:mse_mnist}
\end{figure}

\newpage

\section{ BiGAN results}
Results from Efficient GAN-Based anomaly detection \cite{EffiGAN}. Comparison of BiGAN, AnoGAN and VAE for anomaly detection on MNIST digits
\begin{figure}[!htb]
    \centering
    \includegraphics[width=.7\textwidth]{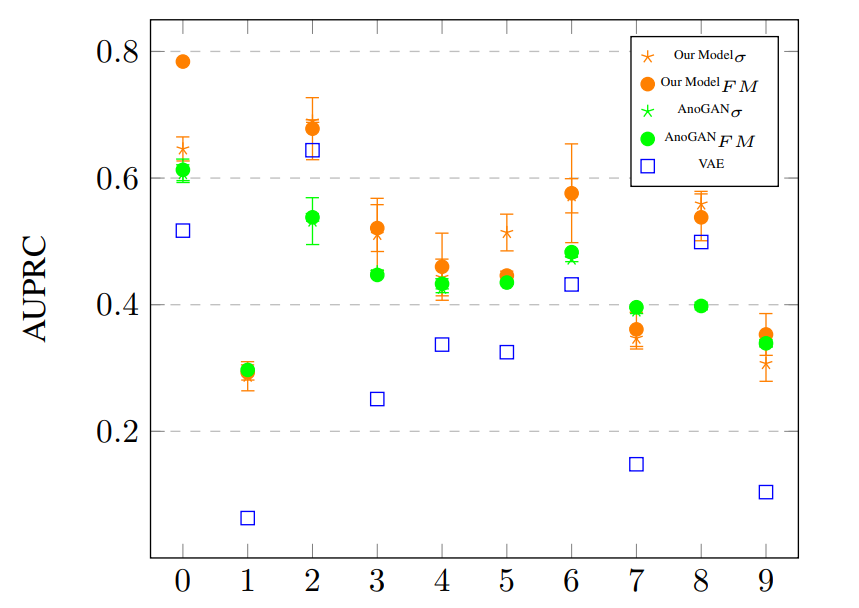}
    \caption{Digit designated as anomalous class, figure from \cite{EffiGAN}}
\label{fig:effi}
\end{figure}
\\
Results in the table of section 5.3 are taken from Figure \ref{fig:effi} whereas results for VAE are numerically cited in \cite{VAE}.
\newpage
\section{GAN Architectures use case 2}

\begin{figure}[H]
\subfigure[Encoder model]{\includegraphics[width=.32\linewidth]{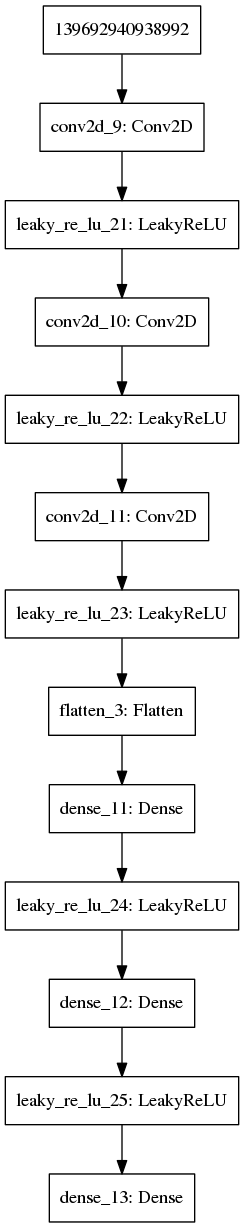}}\label{fig:awesome_image1}
\subfigure[Generator model]{\includegraphics[width=.32\linewidth]{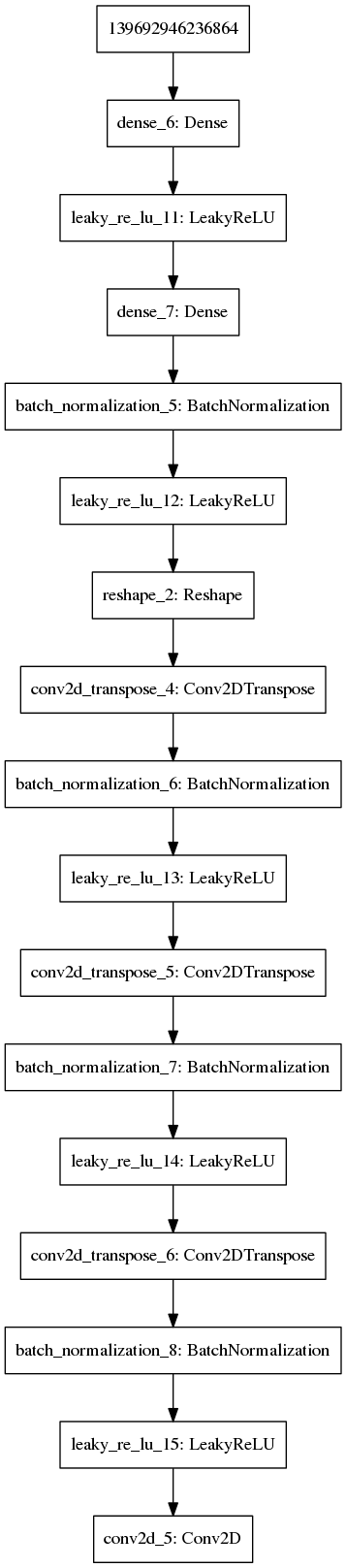}}\label{fig:awesome_image2}
\subfigure[discriminator model]{\includegraphics[width=.32\linewidth]{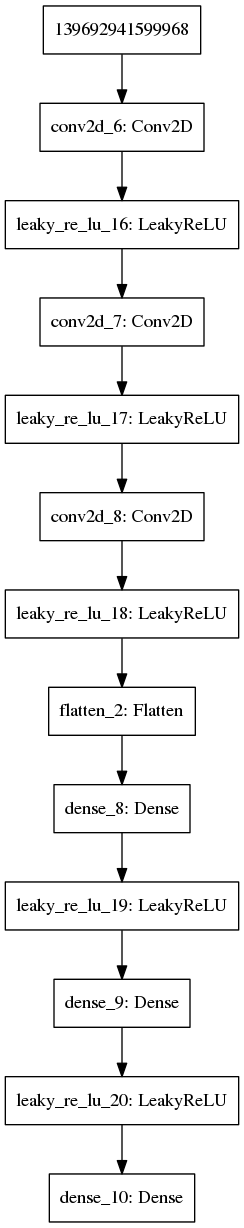}}\label{fig:awesome_image3}
\end{figure}

\section{Implementation details of WGAN}
Here we present some details about the implementation of the Wasserstein loss function
\subsubsection{WGAN loss}\hfill\\
\par The GAN loss function for the generator writes as:
\begin{equation}
    -\frac{1}{m} \sum\limits_{i=1}^m f_w(g_{\theta}(z^{(i)}))
\end{equation}

The discriminator loss is:

\begin{equation}
     \frac{1}{m} \sum\limits_{i=1}^m f_w(x^{(i)}) - \frac{1}{m} \sum\limits_{i=1}^m f_w(g_{\theta}(z^{(i)}))
\end{equation}

These loss functions are not available in keras and need to be defined from scratch. We define an appropriate loss in python as follow:
%
%
\begin{python}
import keras.backend as K

def wasserstein(y_true,y_pred):
    return K.mean(y_true*y_pred)
\end{python}
This function is to be fed to the chosen keras optimizer RMSprop:
\begin{python}
optimizer_G = RMSprop(lr=0.00005)
optimizer_D = RMSprop(lr=0.00005)
\end{python}

The loss function compares at each back-propagation, the true value of labels of the minibatch (-1 if the corresponding data comes from the generator , 1 if the data comes from the critic) with the actual "score" given by the critic (which can differ from -1 and 1).  \\ If y\_true and y\_pred are m-dimensional vectors with m the length of the minibatch of data, then:
\begin{equation}
\texttt{wasserstein(y\_true,y\_pred)} = \frac{1}{m} \sum\limits_{i=1}^m (-1)^{s(i)}f_w(y^{(i)})
\end{equation}
Where $y^{(i)} = x^{(i)}$ if it is a true data and $y^{(i)} = g_{\theta}(z^{(i)})$ if the data was generated. $s(i) = 0$ if the $i^{th}$ example is true and $s(i) = 1$ if it was generated.
\\ As a result, performing back-propagation on a minibatch of true samples leads on the one hand to the gradient descent along the first part of the critic loss $\frac{1}{m} \sum\limits_{i=1}^m f_w(x^{(i)})$, a second back-propagation on a minibatch of fake samples leads on the other hand to the gradient descent along the second part of the critic loss $- \frac{1}{m} \sum\limits_{i=1}^m f_w(g_{\theta}(z^{(i)}))$. In a nutshell, performing 2 back-propagation on the critic weights on true samples and fake ones leads to the descent along the gradient calculated in the pseudo-code above $\nabla_{w}[ \frac{1}{m} \sum\limits_{i=1}^m f_w(x^{(i)}) - \frac{1}{m} \sum\limits_{i=1}^m f_w(g_{\theta}(z^{(i)}))]$. \\
Actually, the Wasserstein function as it is defined in python can also be used to compute the gradient descent for the generator. Performing one gradient descent on a minibatch for fake samples, the Wassertein python loss becomes $\sum\limits_{i=1}^m (-1)f_w(g_{\theta}(z^{(i)}))$ and the corresponding gradient is the same one as in the pseudo code: -$\nabla_{w} \frac{1}{m} \sum\limits_{i=1}^m f_w(g_{\theta}(z^{(i)}))$. \\
This is the explanation for the standard Wasserstein loss. 
\subsubsection{Penalization implementation} \hfill\\
When gradient penalty is performed over the objective function of the Wasserstein GAN, gradients of the discriminator are defined everywhere in $\mathcal{X}$ and the gradient $\nabla_x D_{\alpha}$ are available. Unfortunately, enforcing this constraint everywhere is intractable, and the implementation found only involves the calculation of gradients between randomly sampled points between a batch of elements in $\mathcal{X}$. At the end, the gradients of the discriminator with respect to its input converges toward 1 during the training. 

\section{Additional information}
Here are some plots that have been useful to compute AUPRC for MNIST and UCI data:
\begin{figure}[!htbp]
\centering
\subfigure{\includegraphics[height=4cm,width=.45\linewidth]{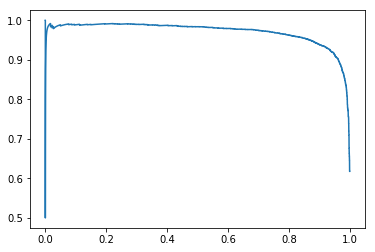}}
\subfigure{\includegraphics[height=4cm,width=.45\linewidth]{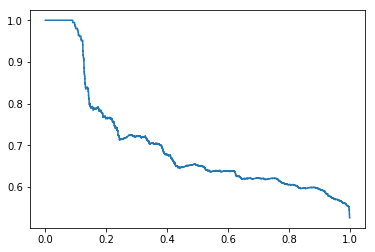}}
\caption[Two numerical solutions]{Left: Precision-Recall curve for digit 0 considered abnormal; Right: Precision-Recall curve for "laying" activity considered abnormal }
\label{fig:PR}
\end{figure}
\\
These curves and their corresponding AUPRC have been computed with scikit-learn python library. 
\end{document}